\documentclass[lettersize,journal]{IEEEtran}
\usepackage{times}
\usepackage{epsfig}
\usepackage{graphicx}
\usepackage{amssymb}
\usepackage{amsmath,amsfonts}
\usepackage{array}
\usepackage[caption=false,font=normalsize,labelfont=sf,textfont=sf]{subfig}
\usepackage{textcomp}
\usepackage{stfloats}
\usepackage{url}
\usepackage{verbatim}
\usepackage{graphicx}
\usepackage{booktabs}
\usepackage{multirow}
\usepackage{fancyhdr}
\usepackage[inline]{enumitem}
\usepackage{algorithm}
\usepackage{algpseudocode}
\usepackage{xcolor}
\usepackage{color}
\usepackage{tabularray}
\usepackage[colorlinks = true,
            urlcolor  = magenta,
            citecolor = black,
            ]{hyperref}

\def\BibTeX{{\rm B\kern-.05em{\sc i\kern-.025em b}\kern-.08em
    T\kern-.1667em\lower.7ex\hbox{E}\kern-.125emX}}
\usepackage{balance}
\begin{document}
\title{A DeNoising FPN with Transformer R-CNN for Tiny Object Detection}
\author{Hou-I Liu, Yu-Wen Tseng, Kai-Cheng Chang, Pin-Jyun Wang, Hong-Han Shuai, Member, IEEE, Wen-Huang Cheng, Senior Member, IEEE
\thanks{Hou-I Liu, Hong-Han Shuai, and Pin-Jyun Wang are with the Department of Electronics and Electrical Engineering, National Yang Ming Chiao Tung University, Hsinchu, 300 Taiwan (email: {k39967.c, hhshuai, whc666777.sc06}@nycu.edu.tw).
Yu-Wen Tseng and Kai-Cheng Chang are with the Institute of Electronics, National Yang Ming Chiao Tung University, Hsinchu, 300 Taiwan (email: {ywt0226.ee10, kaicheng026.ee10}@nycu.edu.tw).
Wen-Huang Cheng is with the Department of Computer Science and Information Engineering, National Taiwan University (NTU), Taipei, 106 Taiwan (email: wenhuang@csie.ntu.edu.tw).}}


\markboth{Journal of \LaTeX\ Class Files,~Vol.~18, No.~9, September~2020}%
{How to Use the IEEEtran \LaTeX \ Templates}

\maketitle

\begin{abstract}
Despite notable advancements in the field of computer vision, the precise detection of tiny objects continues to pose a significant challenge, largely owing to the minuscule pixel representation allocated to these objects in imagery data. This challenge resonates profoundly in the domain of geoscience and remote sensing, where high-fidelity detection of tiny objects can facilitate a myriad of applications ranging from urban planning to environmental monitoring. In this paper, we propose a new framework, namely, DeNoising FPN with Trans R-CNN (DNTR), to improve the performance of tiny object detection. DNTR consists of an easy plug-in design, DeNoising FPN (DN-FPN), and an effective Transformer-based detector, Trans R-CNN. Specifically, feature fusion in the feature pyramid network is important for detecting multiscale objects. However, noisy features may be produced during the fusion process since there is no regularization between the features of different scales. Therefore, we introduce a DN-FPN module that utilizes contrastive learning to suppress noise in each level’s features in the top-down path of FPN. Second, based on the two-stage framework, we replace the obsolete R-CNN detector with a novel Trans R-CNN detector to focus on the representation of tiny objects with self-attention. Experimental results manifest that our DNTR outperforms the baselines by at least 17.4\% in terms of $\text{AP}_{vt}$ on the AI-TOD dataset and 9.6\% in terms of AP on the VisDrone dataset, respectively. Our code will be available at \href{https://github.com/hoiliu-0801/DNTR}{https://github.com/hoiliu-0801/DNTR}.

\end{abstract}
\begin{IEEEkeywords}
Aerial image, tiny object detection, contrastive learning, noise reduction, transformer-based detector.  
\end{IEEEkeywords}

\section{Introduction}
\label{sec:intro}

\IEEEPARstart{W}{ith} the remarkable advancement of deep learning, object detection has witnessed significant progress over the past decade. Object detection techniques can generally be divided into two main categories: two-stage detectors~\cite{detectors,frcnn,crcnn} and one-stage detectors~\cite{retinanet,fcos,foveabox}. While two-stage detectors often yield higher accuracy, one-stage detectors boast faster inference speeds, making them more suitable for real-time detection. However, most previous works focus on general object detection, while the methods for detecting tiny objects have not been explored well. In fact, tiny objects are pervasive in aerial images and are important to a variety of applications, \textit{e.g.}, traffic monitoring~\cite{rsod}, surveillance~\cite{surveillance}, and sea rescue~\cite{rescue}. 

To this end, we focus on tiny object detection whose order of magnitude is generally smaller than that of small object detection. Following~\cite{stdnet,aitod}, the definition of a tiny object is an object smaller than $16 \times 16$ pixels, whereas the small objects in COCO~\cite{coco} are smaller than $32 \times 32$ pixels. The tiny size makes detection challenging. For instance, the widely-used CNN model~\cite{resnet} reduces the input resolution by a factor of 32, resulting in an insufficient representation of tiny objects~\cite{IPG-net,querydet,rjfpn}. 

\begin{figure*}[htbp] 
\centering 
                \includegraphics[width=1.5\columnwidth]{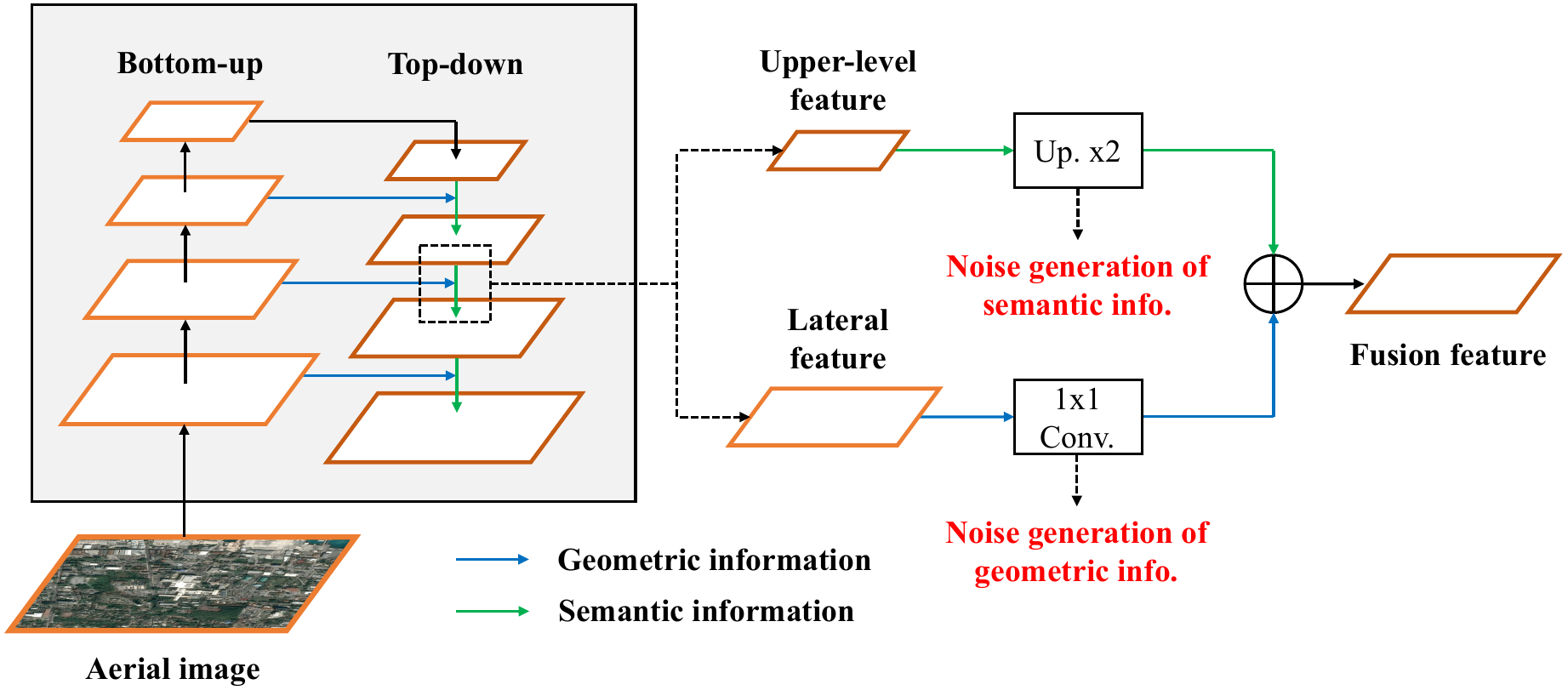}
                \caption{Conventional FPN structure. Following FPN~\cite{fpn}, the fusion features are combined with the lateral features and upper-level features. The main purpose is to aggregate the geometric and semantic information from the low-resolution and high-resolution features to garner better multiscale features. However, the channel reduction (1x1 Conv.) and upsampling (Up. x2) cause noise and damage the geometric and semantic information in FPN, respectively. Note that the Conv. and Up. denote the convolution layer and the upsampling operation.}
                \label{fig:fpn}
\end{figure*}

Traditionally, to accurately detect objects at different scales, an FPN~\cite{fpn} aggregates multiscale features by a top-down path to incorporate both geometric and semantic information, as depicted in Fig.~\ref{fig:fpn}. Specifically, the lateral features from the bottom-up path provide geometric information, while the upper-level features from the top-down path offer semantic information. Unfortunately, upsampling methods, such as bilinear interpolation and nearest neighbor methods, cause redundant information, while channel reduction leads to information loss in the features. In these cases, The distortion of features is considered noise generation in the fusion process of FPN~\cite{IPG-net,eff-fpn,dual,Trifpn}. As shown in Fig.~\ref{fig:fpn}, the channel reduction may introduce noise in the geometric information, whereas the upsampling operation may produce noise in the semantic information. Consequently, the noisy feature maps of FPN adversely affect the detection of tiny objects since the low-resolution features are more noise-sensitive.

A recent line of studies~\cite{dual,Trifpn,TGRS_FSA} focuses on leveraging an additional module to enhance feature fusion while alleviating redundant feature production.
For instance,~\cite{Trifpn} builds a feature feedback module to capture better communication between bottom-up and top-down features.~\cite{TGRS_FSA} introduces an $FA^2M$ module to learn the pixel-spatial information to align the multilevel features. 
Nonetheless, these methods require an extra module to improve their performance, leading to a cumbersome model. Furthermore, some works aim to recall tiny objects from the background based on their properties. For example,~\cite{nwd,nwd2,rfla} design new label assignment methods based on Gaussian to replace the IoU metric since the IoU metric is too sensitive to tiny objects.~\cite{dot} considers the center point distance and the average size of the dataset in the new loss function for RPN and NMS. Although the above methods have achieved certain improvements, they only focus on enhancing the performance of tiny objects to offset the impact of noise. None of them suppress the noise at its source, i.e., the FPN fusion process. Hence, we pioneer in minimizing the FPN's noise problem to boost the performance of tiny objects.

\begin{figure*}[htbp] 
\centering 
                \includegraphics[width=1.5\columnwidth]{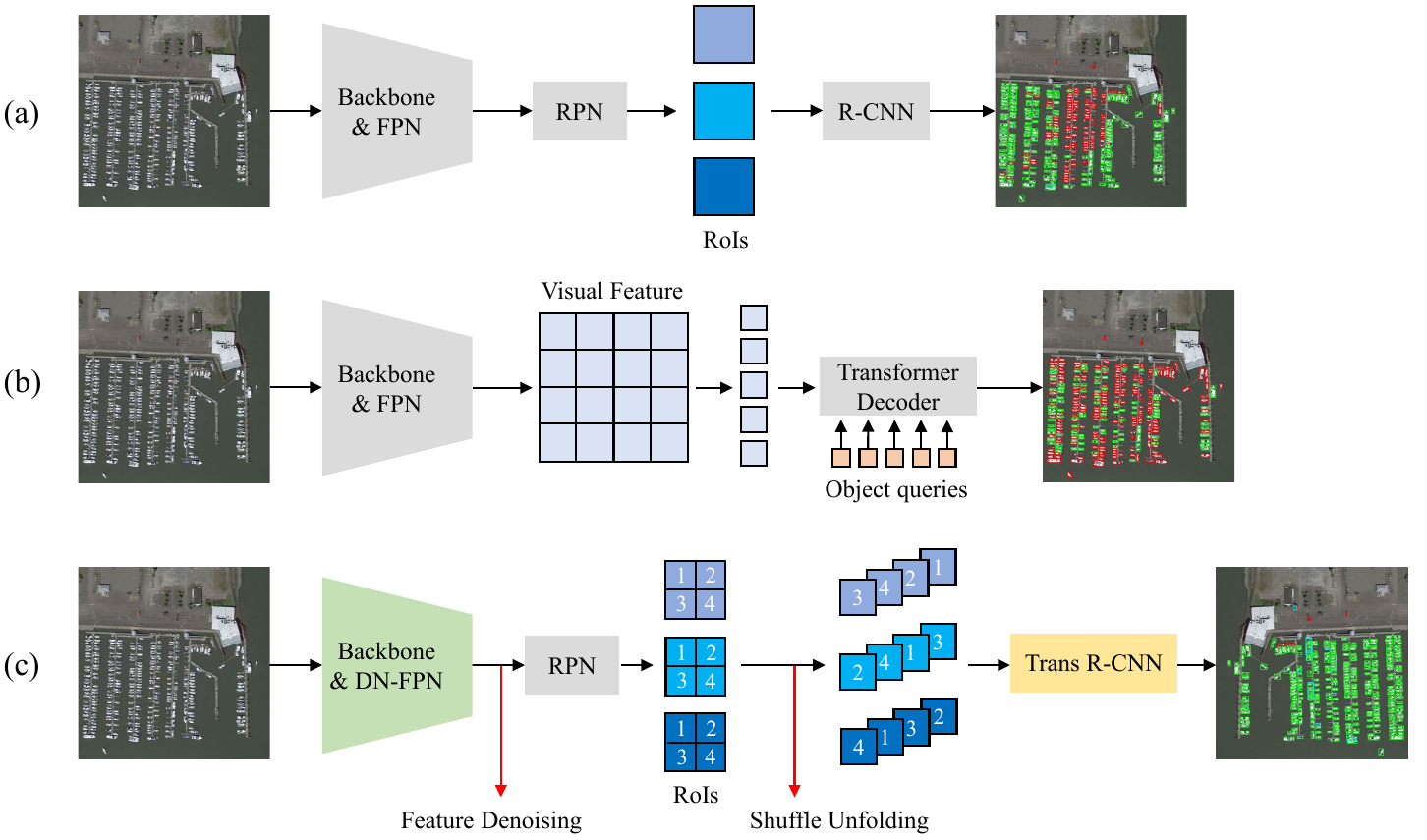}
                \caption{Comparison of different object detection frameworks. (a) CNN-based two-stage model~\cite{detectors} generates the RoIs (Three different shades of blue) by RPN and applies R-CNN head to predict objects. (b) DETR-like model~\cite{detr} flattens the visual feature into image patches and passes them through the transformer decoder via cross-attention to transform the object queries into the final bounding boxes. (c) Our DNTR extracts less noisy multiscale features by the DN-FPN module. Subsequently, the shuffle unfolding and Trans R-CNN are employed to capture local and global information within an RoI, resulting in better detection outcomes for tiny objects.}
                \label{fig:pipeline}
\end{figure*}

In this work, we propose De-Noising FPN (DN-FPN) to leverage contrastive learning to reduce the FPN's noise problem. The objective of DN-FPN is to uphold both geometric and semantic information within the fusion features, as illustrated in Fig.~\ref{fig:fpn}. Specifically, we first use geometric and semantic encoders to extract the representations from different layers of FPN's features. To supervise the encoders, we propose a novel contrastive loss to learn the geometric and semantic relation of FPN's features according to their inherent characteristics. As such, the fusion features maintain the equivalent geometric information with the lateral feature (from the bottom-up path) and retain the equivalent semantic information with the upper-level feature (from the top-down path) after contrastive learning. Consequently, the noise generated from channel reduction and upsampling is reduced\footnote{It is worth noting that the DN-FPN module only applies in the training stage, so it can be incorporated into other FPN-based models without adding additional FLOPs and parameters during inference (see Table~\ref{tab:fpn}).}.

Moreover, two-stage detectors are renowned for their precision in localization and recognition~\cite{detectors,frcnn,rcnn}. However, they often struggle with the nuanced detection of tiny objects due to insufficient pixel representation of tiny objects~\cite{TGRS_track, TGRS_structure}. To address this, we introduce Trans R-CNN, a two-stage novel detector that integrates the strengths of vision transformers~\cite{vit}, fostering the extraction of rich, long-range dependencies. 
Specifically, Trans R-CNN first deconstructs each proposal into a series of patch tokens. These tokens, representing distinct facets of tiny objects, offer a more granular and effective representation than traditional, whole-proposal approaches. Subsequently, we propose a shuffle unfolding algorithm aimed at aggregating surrounding tokens into unfolded tokens to enable the assimilation of local information, which serves as more discriminative representations for tiny objects. 
  
Our contribution in this space, represented by the DN-FPN module and Trans R-CNN detector, distinguishes itself by proficiently extracting multiscale features devoid of noise interference, positioning itself as a more precise tool optimized for tiny object detection. As illustrated in Fig.~\ref{fig:pipeline}, we compare DNTR with traditional frameworks to underline its superiority. While R-CNN-based detectors excel in localizing potential object regions, they fall short in global information capture (Fig.~\ref{fig:pipeline}(a)). Moreover, DETR-like systems are prone to a high rate of false positives due to excessive background responses stemming from the holistic handling of visual features (Fig.~\ref{fig:pipeline}(b)), as indicated by~\cite{DQ-detr}. In contrast, our DNTR approach leverages the DN-FPN module to emphasize foreground elements through the creation of denoising multiscale features (Fig.~\ref{fig:pipeline}(c)). The subsequent integration of RPN and RoIAlign~\cite{mrcnn} techniques\footnote{We use region proposal network (RPN) to generate region proposals, and employ RoIAlign to transform region proposals into regions of interest (RoIs).}, followed by shuffle unfolding, facilitates a richer local information pool within the RoIs by synergizing with neighboring tokens. The Trans R-CNN detector further accentuates global information capture within the RoIs by the proposed Mask Transformer Encoder (MTE), thereby enhancing the discernibility of tiny objects. 

The contributions are summarized as follows:

\begin{itemize}
    \item We present a groundbreaking method for object detection in aerial imagery, substantiated by standout performance on the AI-TOD and VisDrone datasets, where our DNTR model surpasses prevailing state-of-the-art models with AP scores of 26.2 and 33.1, respectively.
    \item We unveil the DN-FPN, designed through geometric-semantic contrastive learning, a technique pioneered to mitigate the impact of redundant features frequently encountered in FPN-based noisy feature fusion.
    \item Our development of the Trans R-CNN epitomizes innovation in feature discernment, utilizing a novel shuffle unfolding technique, mask transformer encoder, and task token selection mechanism to foster a more holistic and detailed feature extraction, enhancing both local and global discriminative features.
\end{itemize}

\section{Related Works}
\label{sec:op}
\noindent\textbf{Resolution Enhancement and Complex Network Architectures.}
Since small objects are blurry and lack detailed information, several works delve into increasing the resolution of images based on super-resolution (SR) technologies~\cite{ESRTMDet, TGRS_superyolo}. 
For instance,~\cite{ESRTMDet} creates a super-resolution branch to restore the high-resolution image and applies feature alignment loss to preserve the high-frequency texture information. Moreover,~\cite{TGRS_superyolo} fuses the infrared and RGB image features to enhance the surface details and performs super-resolution to ameliorate the quality of small objects in remote sensing images. On the other hand, some methods present a complex network that can incorporate both high-level and low-level features simultaneously. For instance,~\cite{hrdnet} proposes a multi-depth image pyramid for detecting small objects based on~\cite{hrnet}. 
~\cite{DenseGL} employs the involution and self-attention to the backbone to extract spatial features. However, these methods cause an extra computation cost due to using a higher-resolution image or a more complex model to improve the performance of small object detection.

\noindent\textbf{Advancements in Feature Pyramid Networks (FPNs).}    
Considering the multiscale concept provides great benefits to object detection, many studies aim to optimize the network based on FPN~\cite{fpn}. For example,~\cite{eff-fpn} proposes an efficient fusion factor to re-weight feature maps during the fusion process since semantic information of small objects can be attenuated by an effective fusion factor to reduce the information from the top layer. Moreover,~\cite{TGRS_LAP} rotates the high-level semantic feature of FPN into four different degrees, concatenating them along the channel dimension and then passing through the convolution layer. In this way, the interactions between different perspectives of high-level semantic features can further reinforce the global semantic information.
~\cite{rjfpn} generates the FPN features by applying adaptive hierarchical upsampling, which can provide great semantic compensation to low-level features, thereby preventing the dilution disturbance problem caused by the FPN fusion. Nevertheless, the above variant FPN methods also require extra modules or more powerful inputs, such as additional high-resolution images or additional modules.
In contrast, the proposed DN-FPN module is only used in the training phase and raises performance significantly without increasing image resolution.

\noindent\textbf{Coarse-to-Fine Image Partitioning in Drone Imagery Detection.}
Different from general object detection, drone imagery detection~\cite{UFPMP,amrnet,clusterdet,density-crop, TGRS_CFA} follows a coarse-to-fine framework to obtain more precise detection results, i.e., partitioning the input images into small patches and then using resized patches as the input for detection networks. For instance,~\cite{TGRS_CFA} splits images into equal sizes with overlapping parts to eliminate the discontinuity of objects at the boundaries. Moreover,~\cite{density-crop} introduces a density-guided training method to obtain the density areas, which are then zoomed in to detect small and crowded objects better.~\cite{UFPMP,amrnet} utilize mosaic packing to merge multiscale images, increasing the foreground ratio and speeding up the inference process. 
Although the performance gain is significant, these coarse-to-fine frameworks require additional time consumption in the post-processing fusion and also produce many partial objects.


\noindent\textbf{Developments in Loss Function Approaches for Tiny Object Detection.}
In tiny object detection, general IoU metrics are too sensitive to slight offsets between predicted bounding boxes and ground truth data. Therefore, DotD~\cite{dot} concentrates on the relative distance between center points by taking the average size of all objects in a dataset into account as a new metric. NWD~\cite{nwd} further models bounding boxes as 2D Gaussian distributions and uses the normalized Wasserstein distance to alleviate the sensitivity of the location deviation of tiny objects. Based on NWD~\cite{nwd}, instead of setting a fixed threshold, NWD-RKA~\cite{nwd2} samples the top-K high-quality regions to avoid assigning a large number of negative labels.~\cite{TGRS_fmssd} introduces an area-weighted loss to strengthen the learning of small objects, where the weight increases as size decreases.

\noindent\textbf{R-CNN for Small Object Detection.} Drawing inspiration from R-CNN~\cite{rcnn}, \cite{rcnn_s1} underscores that the proposals generated by RPN are too large for small objects, resulting in poor efficacy. To this end, they employ the smaller anchor boxes tailored for detecting smaller objects, aligning with the dataset's statistical characteristics. Moreover, \cite{rcnn_s2} observes that the small objects are easily mismatched with the different scale anchors since the small objects have fewer samples in the training data. Thus, they adopt a strategy of copy-pasting the small objects onto images without overlap to oversample and augment the quantity of smaller objects. Furthermore, \cite{rcnn_s3} focuses on enlarging the proposals via super-resolution so as to alleviate the poor resolution of small objects.

\noindent\textbf{Transformer Models Leveraging Self-Attention in Object Detection.} Vision Transformer (ViT)~\cite{vit} introduces a pivotal shift in object detection, where images are broken down into non-overlapping tokens to be analyzed through transformer layers for classification tasks. Despite its innovation, the method faces issues due to weakened relationships between adjacent tokens, a problem addressed by~\cite{t2t} through the utilization of unfolded operations to enhance local relations. Building on this, DETR~\cite{detr} initiates an end-to-end object detection framework that relies on transformer mechanics to identify object boundaries, although it struggles with detecting small objects and has long training periods. To approach this issue,~\cite{ddetr} introduces multiscale features into a deformable attention module, enhancing small object detection and accelerating convergence. Further improvements are achieved by~\cite{dino,dndetr} with the deployment of contrastive denoising training methods, significantly enhancing performance on the COCO dataset~\cite{coco}.

Although transformer models can better leverage self-attention, they are still fatally flawed in handling tiny objects. 
There is usually an imbalance between the foreground and background of remote-sensing images due to the tiny size of objects and their small proportion of the image.
Additionally, tiny objects are blurry and can easily be confused with the background, as DETR-like methods use the entire feature map as input, resulting in high false positives (FP)~\cite{DQ-detr}.
More precisely, the self-attention layers mostly perform similarity calculations with the background, which is inefficient and confusing. 
To overcome this obstacle, our Trans R-CNN combines the advantages of RPN and self-attention, which can better capture the local and global information within the region proposal.

\section{Method}
\label{sec:Methodology}

Due to the lack of pixel representation of tiny objects, their representations are more sensitive to noise. Additionally, integrating the local and global information from the surrounding pixels enhances the representation of tiny objects. To this end, we present an effective detection framework, namely DNTR, for tiny object detection. The guiding concept of our DNTR is to leverage enhanced geometric and semantic relations, effectively capturing the local and global information within an RoI feature, thereby improving the performance of tiny objects. As such, two goals can be achieved: 
\begin{enumerate*}[label=(\arabic*)]
\item The noise of FPN features introduced during the fusion can be reduced by preserving pure geometric and semantic information (DN-FPN).
\item The RoI features can be better leveraged to capture local and global information, resulting in better detection (Trans R-CNN).
\end{enumerate*} 

\begin{figure*}[t] 
\centering 
                \includegraphics[width=1.7\columnwidth]{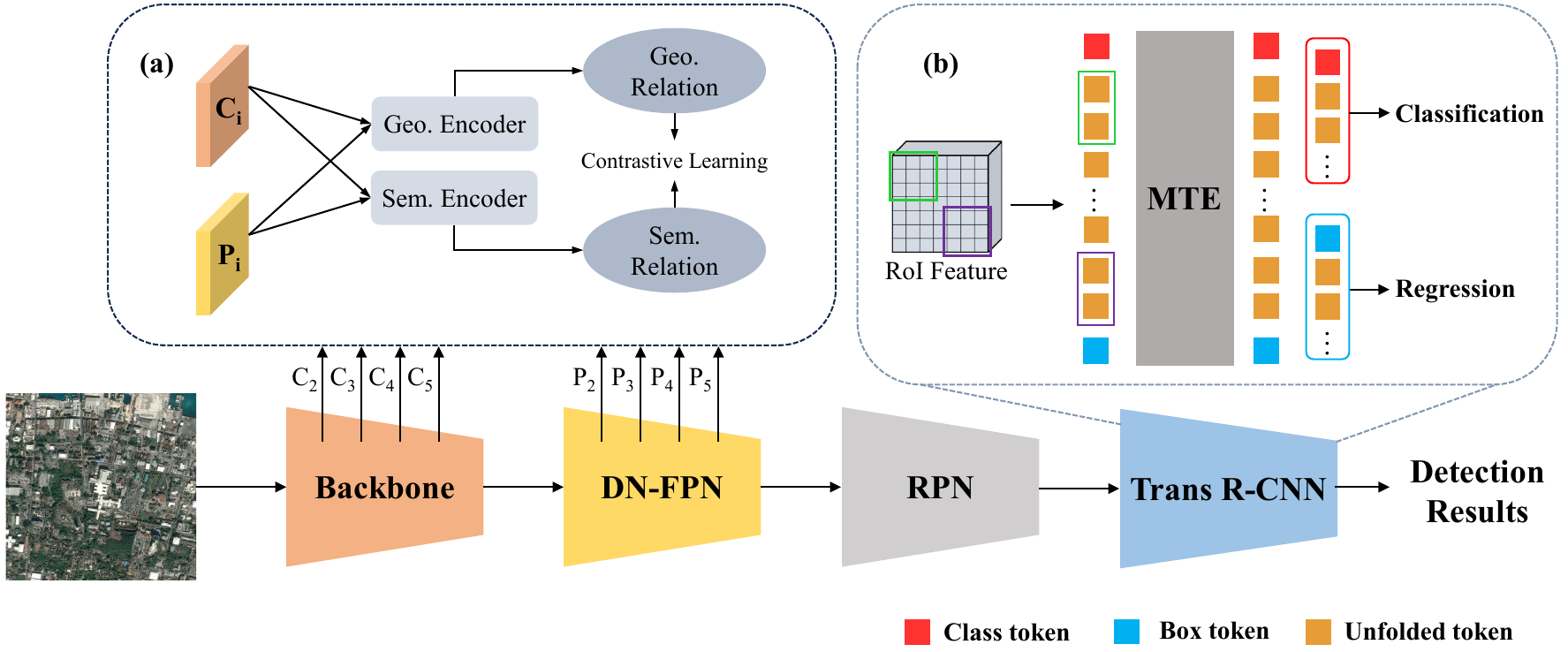}
                \caption{Overall architecture of DNTR. We use $C_{i}$ and $P_{i}$ to represent the multiscale features from the backbone and DN-FPN (Eq.~\ref{eq:1}), respectively, where i denotes the levels of the multiscale feature. (a) DN-FPN module, which suppresses redundant features by geometric-semantic contrastive learning. Note that the Geo. and Sem. represent the geometric and semantic, respectively. (b) Trans R-CNN head, which aims to utilize the surrounding information and capture rich long-range dependencies within an RoI.}
                \label{fig:overall}  
\end{figure*}

Fig.~\ref{fig:overall} shows the overall structure of DNTR, where Fig.~\ref{fig:overall} (a) illustrates the proposed DN-FPN, which is an FPN with geometric and semantic encoders optimized by the proposed contrastive loss. This part aims to reduce the noise generated during the FPN fusion process. Afterward, the FPN features undergo the RPN network and the RoIAlign~\cite{mrcnn} to obtain the RoIs. Fig.~\ref{fig:overall} (b) presents the proposed innovative detector, Trans R-CNN, designed to enhance local relations and capture more global information within an RoI. 
We will describe the DN-FPN and Trans R-CNN in detail in the following section.
 
\subsection{DN-FPN} 
FPN~\cite{fpn} has greatly improved the development of object detection by merging the features from different levels. The fusion process at the $i$-th level of an FPN can be formulated as:


\begin{equation}\label{eq:1}
P_{i} =
\begin{cases}
Conv(C_{i})+Up(P_{i+1}), i=0,...,L-1\\
Conv(C_{i}), i=L.
\end{cases}
\end{equation}
where $P_{i}$ represents the features in the top-down path, $C_{i}$ represents the features in the bottom-up path, i.e., the multiscale features of ResNet50 backbone\footnote{We use ResNet50 as the backbone of DNTR to compare our results with other state-of-the-art methods easily.} produced by downsampling (referred to as Fig.~\ref{fig:fpn}). Additionally, $L$ denotes the number of the levels of FPN. $Conv(\cdot)$ is the convolution operation for the channel reduction and $Up(\cdot)$ is the upsampling operation. 
As a consensus of FPN, the lateral features contain strong low-level information since the high-resolution feature maps represent the geometric representation. In contrast, the upper-level features have abundant semantic representation due to the deeper feature extraction.


Since the geometric information undergoes distortion (referred to as noise) due to channel reduction, while the semantic information faces similar issues due to upsampling, we formulate the FPN noise problem as the geometric and semantic information loss problems. To solve these problems, we propose De-Noising FPN (DN-FPN), a contrastive learning method that eliminates noise generated by channel reduction $Conv(C_{i})$ and upsampling $Up(P_{i+1})$. DN-FPN aims to preserve the geometric information from the lateral feature $C_{i}$ and the semantic information from the upper-level feature $P_{i+1}$ for eliminating the noise in the fusion feature $P_{i}$.


\noindent \textbf{Geometric and Semantic Representations.}
First, we employ geometric and semantic encoders to extract the geometric and semantic information from the features across each level of the FPN within the minibatch. 
Specifically, we project the lateral feature $C_{i,j}$ into the geometric representation and the semantic representation $\{g^c_{i,j}, s^c_{i,j} \in\mathbb{R}^{256}\}$ by the geometric and semantic encoders, respectively, where $i$ denote the $i$-th level of an FPN and $j$ denotes the index of different samples in a minibatch. Similarly, the upper-level feature $P_{i,j}$ is also projected into the geometric representation and semantic representation $\{g^p_{i,j}, s^p_{i,j} \in\mathbb{R}^{256}\}$ by the geometric and semantic encoders, respectively. It is worth noting that we consider different levels and batches of geometric and semantic representations in the contrastive loss.

\begin{figure*}[htbp] 
    \centering 
    \includegraphics[width=1.8\columnwidth]{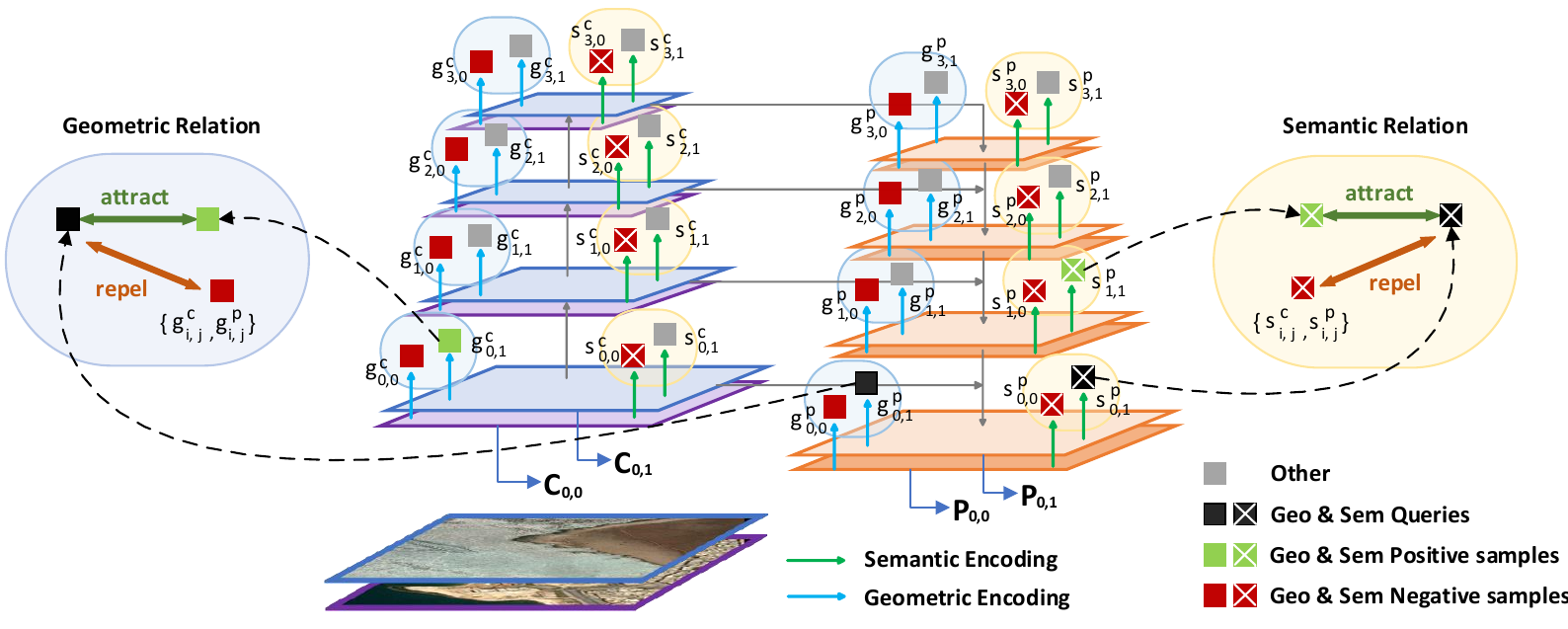}
    \caption{An illustration of geometric and semantic relations of DN-FPN. We set the $g^p_{0,1}$ and $s^p_{0,1}$ as the example queries (black frames) of the geometric and semantic representation of the lowest fusion feature $P_{0,1}$. The geometric and semantic relations of positive and negative samples are shown in the light blue and light yellow oval regions, respectively. It is important to note that the queries, positive and negative samples, are computed independently in the geometric and semantic relations, highlighted in solid and cross frames. In addition, {\color{gray}\textbf{Other}} denotes the representations that are not included in the loss function.}
    \label{fig:cl}
\end{figure*}

Given $L$ levels of an FPN and $B$ images within a minibatch, we decompose the feature maps of FPN into the sets of the geometric representations $G=\{g^c_{i,j}, g^p_{i,j} \mid i=0,...,L;j=0,...,B\}$ and semantic representations $S=\{s^c_{i,j}, s^p_{i,j} \mid i=0,...,L;j=0,...,B\}$, as shown in Fig.~\ref{fig:cl}.
Subsequently, we strive to reduce the distance between the fusion features' geometric information and the lateral features' geometric information, as the lateral feature represents undisturbed geometric information.
Similarly, we bring the semantic information of the fusion feature and upper-level feature closer together. Conversely, we aim to repel the geometric and semantic information of fusion features from different levels and batches, which are unrelated to the fusion features.

Specifically, we utilize the InfoNCE loss~\cite{moco} to learn the better relationship among geometric representations $G$, called geometric relation, and among semantic representations $S$, called semantic relation. Establishing a better relationship between geometric and semantic representations can lead to noiseless feature fusion, making the FPN structure reliable. The positive and negative samples of the geometric and semantic relations are introduced as follows.

\noindent\textbf{Geometric Relation.} As depicted in Fig.~\ref{fig:cl}, the lateral features $C_{i,j}$ are used to provide geometric information to the fusion feature $P_{i,j}$. Ideally, the geometric representations from $C_{i,j}$ and $P_{i,j}$ should be identical. However, the channel reduction damages the top-down features $P_{i,j}$, which causes the unequal geometric information to the lateral features $C_{i,j}$. 

Given the $k$-th level of the top-down path and the $b$-th batch, we regard the geometric representations $g^p_{k,b}$ (query) and $g^c_{k,b}$ (positive sample) as the positive pair since the geometric information should be as similar as possible in the lateral direction. 


Conversely, we define all different levels of geometric representations from different images within the mini-batch as negative samples since they have different geometric information.
The negative sample set for geometric relation ($G^-$) can be denoted by

\begin{equation}\label{eq:2}
  G^- = \{g^c_{i,j}, g^p_{i,j} \mid i=0,...,L, j=0,...,B; j \neq b \}. 
\end{equation}

As such, the geometric loss at the $k$-th level for the $b$-th batch, denoted by $\mathcal{L}_g(\mathbf{g^p_{k,b}})$, is derived by
\begin{equation}\label{eq:3}
     \mathcal{L}_g(\mathbf{g^p_{k,b}})=-log(\dfrac{e^\mathbf{{g^p_{k,b}} \cdot \mathbf{g^c_{k,b}} / \tau}}{{e^{\mathbf{g^p_{k,b}} \cdot \mathbf{g^c_{k,b}} / \tau}}+\sum_{\mathbf{g}\in G^-} e^{\mathbf{g^p_{k,b}} \cdot \mathbf{g} / \tau}}).
\end{equation}
where the temperature $\tau$ is used for controlling how concentrated the features are in the representation space. The overall geometric loss can be calculated as follows.

\begin{equation}\label{eq:4}
    \mathcal{L}_{geo} = \dfrac{1}{L\times B} \sum_{k=0}^L\sum_{b=0}^B \mathcal{L}_{g}(\mathbf{g^p_{k,b}}).
\end{equation}

\noindent\textbf{Semantic Relation.}
Following the concept of FPN, upper-level features propagate rich semantic information to lower-level features of the top-down path, i.e., the fusion feature $P_{i,j}$. Theoretically, $P_{i+1,j}$ and $P_{i,j}$ should have identical semantic representations.  
Nonetheless, the upsampling operation may generate redundant noise, losing the semantic information. In this case, the $P_{i+1,j}$ and $P_{i,j}$ are no longer equivalent. 

Thus, given the $k$-th level of the top-down path and the $b$-th batch, we regard the semantic representation $s^p_{k,b}$ (query) and $s^p_{k+1,b}$ (positive sample) as a positive pair.

By contrast, we regard the semantic representations from different images within the mini-batch as negative samples. 
The negative sample set for semantic relation ($S^-$) can be denoted by
\begin{equation}\label{eq:5}
       S^- = \{s^c _{i,j}, s^p_{i,j} \mid i=0,...,L,j=0,...,B; j\neq b \}. 
\end{equation}

Similarly, the semantic loss at the $k$-th level for the $b$-th batch, denoted by $\mathcal{L}_s(\mathbf{s^p_{k,b}})$, is derived by
\begin{equation}\label{eq:6}
    \mathcal{L}_s(\mathbf{s^p_{k,b}}) = -log(\dfrac{e^\mathbf{{s^p_{k,b}} \cdot \mathbf{s^p_{k+1,b}} / \tau}}{e^\mathbf{{s^p_{k,b}} \cdot \mathbf{s^p_{k+1,b}} / \tau} + \sum_{\mathbf{s}\in S^-} e^\mathbf{{s^p_{k,b}} \cdot \mathbf{s} / \tau}}).
\end{equation}

The overall semantic loss can be calculated as follows.
\begin{equation}\label{eq:7}
    \mathcal{L}_{sem} = \dfrac{1}{(L-1)\times B} \sum_{k=0}^{L-1}\sum_{b=0}^B \mathcal{L}_s(\mathbf{s^p_{k,b}}).
\end{equation}

In summary, Fig.~\ref{fig:cl} illustrates the detailed relation of the geometric and semantic representations.
With $L_{geo}$ and $L_{sem}$, the fusion feature preserves consistency with the information source (lateral and upper-level features). As such, tiny objects can be better detected by features generated by DN-FPN.

\subsection{Trans R-CNN}
\begin{figure}[htbp] 
\centering 
        \includegraphics[width=0.85\columnwidth]{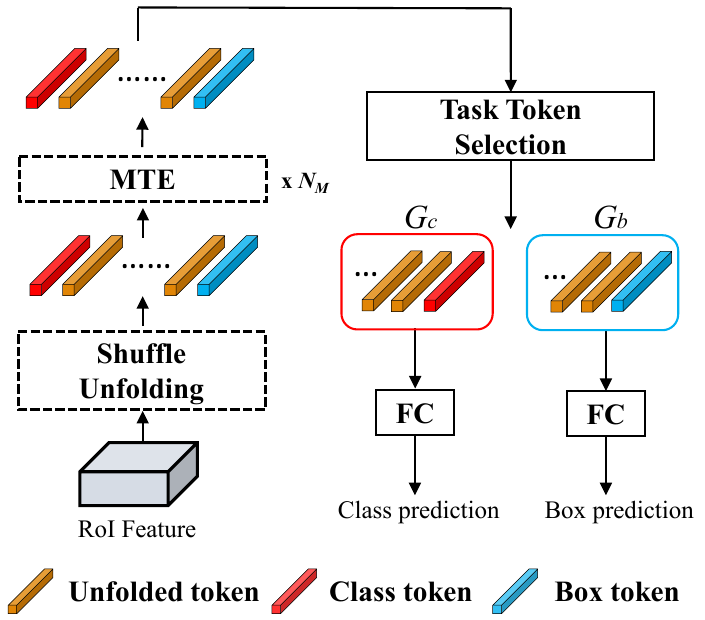}
        \caption{The structure of Trans R-CNN, which is composed of shuffle unfolding, a mask transformer encoder (MTE), and a task token selection mechanism. Please note that the $G_c$ and $G_b$ denote class-related and box-related groups.}
        \label{fig:t2t}
\end{figure}

After DN-FPN obtains enhanced multiscale features, the next step is to better utilize these features for tiny object detection. According to the past two-stage detection method, R-CNN~\cite{rcnn} provides a good way to generate the RoI features for object detection, whereas the long-range dependency is difficult to capture via fully-connected layers. Therefore, we introduce an effective detector, Trans R-CNN, which includes the shuffle unfolding mechanism, the mask transformer encoder, and the task token selection mechanism, as shown in Fig.~\ref{fig:t2t}. The goal of Trans R-CNN is to enhance the representations of tiny objects, as well as to increase global information about those representations.
Thus, we propose the shuffle unfolding to increase the diversity of the tiny object representations and the mask transformer encoder to integrate masked self-attention layers to obtain global information. To expand the benefits of the above methods, the task token selection mechanism is used to better handle the task-specific tokens for the detection head. 

\begin{figure}[htbp] 
\centering 
        \includegraphics[scale=0.5]{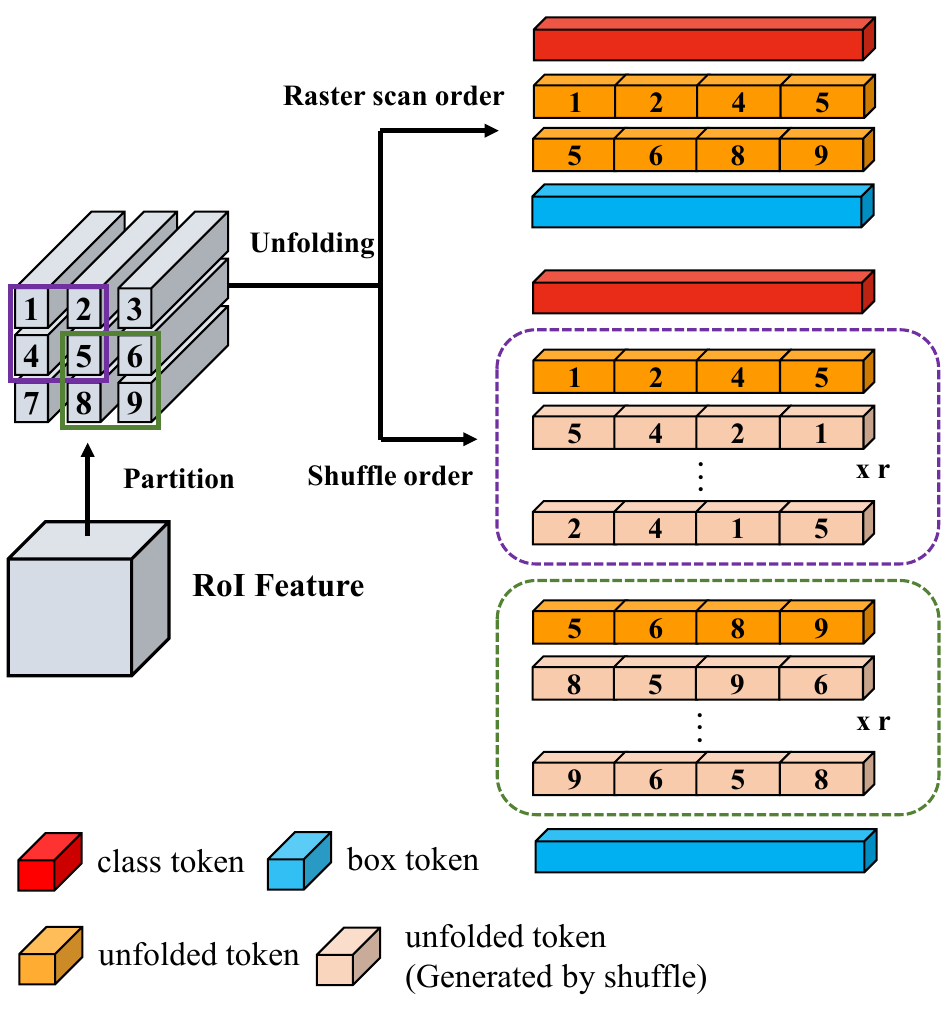}
        \caption{Comparison between raster scan order and shuffle order.}
        \label{fig:shuffle}
\end{figure}

\noindent\textbf{Shuffle Unfolding.}
Fig.~\ref{fig:shuffle} illustrates the proposed shuffle unfolding. This algorithm aims to oversample neighboring patch tokens, amassing them through an unfold operation that preserves intricate spatial details, a strategy pivotal in tiny object detection.

After generating the RoI features by the RPN and RoIAlign, we divide each RoI feature into tokens equally with overlaps. This ensures that each token preserves the continuous local information with the surrounding tokens. Afterward, we concatenate the tokens within the $K\times K$ sliding window as an unfolded token to better leverage the spatial information. For the concatenated order, since using the raster scan order~\cite{t2t} only may limit the feature variety, we propose shuffle order to increase the diversity of the feature. The shuffle order oversamples the combination of unfolded tokens by randomly concatenating them. As the unfolded tokens are concatenated from surrounding tokens, increasing feature diversity can be regarded as enhancing local information.
    
Specifically, the raster scan order only contains the same combination of the unfolded tokens. In contrast, in shuffle order, the combination of the unfolded tokens is proportional to the size of the sliding window, which can exceed the raster scan order by $K^2!$ times. Nevertheless, due to the complexity, we use the oversampling rate $r$ to control the numbers of the generated unfolded tokens. For example, $r=4$ represents $4\times$ token numbers, and another 3 unfolded tokens (light orange) are generated by shuffling from the original one (orange). Let $t_{i}\in R^D$ denote the $i$-th unfolded token with dimension $D$. The unfolded token sequence $T_{unfold} = \{t_{i} | i=1, 2,..., r\}$ can maintain the spatial correlation (tokens' order) within an RoI and get rich spatial information through varying token combinations. Note that the order of the unfolded tokens is fixed after the shuffle unfolding operation. The following operations, such as multi-head self-attention (MSA) layers and fully-connected (FC) layers, maintain this order, thereby preserving the spatial relationship of the unfolded tokens.

\noindent\textbf{Mask Transformer Encoder.}
Equipped with the unfolded token sequence, we further design a Mask Transformer Encoder (MTE) to capture more global information within an RoI, which is an effective encoder consisting of two masked self-attention layers. First, before the attention mechanism, we concatenate the unfolded token sequence $T_{unfold}$ with a class token ($t_{cls}\in R^D$) and a box token ($t_{box}\in R^D$) to form a local token sequence $T_l$:

\begin{equation}
    \label{eq:localseq}
    T_l = \{t_{cls},T_{unfold},t_{box}\}.
\end{equation}

Note that $t_{cls}$ and $t_{box}$ are two learnable tokens representing the task-specific token for classification and regression head, respectively. Second, in order to obtain global information, the MTE transforms the local token sequence $T_l$ into the global token sequence $T_g$ by the multi-head self-attention (MSA) layers. Specifically, $T_{unfold}$ communicates with $t_{cls}$ and $t_{box}$ via MTE to form global token sequence $T_g$, formulated as:

\begin{equation}
    \label{eq:msa}
    \begin{split}
        T_g = MTE(T_l) &= MSA^{*}_{\times N_{M}}(T_l) \\
        &=\{t_{cls}^{'},T_{unfold}^{'},t_{box}^{'}\},
    \end{split}
\end{equation}
where * indicates that a mask is applied to ignore the computation of attention scores between $t_{cls}$ and $t_{box}$ because they are irrelevant to each other. $N_{M}$ denotes the number of MSA layers.
With the mask applied, the self-attention layers are capable of simultaneously processing class and box information.

\noindent\textbf{Task Token Selection.}
After obtaining a global token sequence $T_g$, we design a task token selection algorithm for evenly separating the global unfolded token sequence into a class-related group $G_c$ or a box-related group $G_b$. The selected mechanism is based on the importance of the unfolded token to the class token $t_{cls}^{'}$ and the box token $t_{box}^{'}$. Specifically, for the $i$-th unfolded token,
we calculate the attention score $(\alpha_{cls}^i,\alpha_{box}^i \in \mathbb{R})$ as follows.

\begin{equation}
    \label{eq:softmax1}
    \alpha_{cls}^i=softmax(\frac{Q(t_{cls}^{'})\cdot (K(t_{i}^{'}))^T}{\sqrt{d_{k}}}),
\end{equation}
\begin{equation}
    \label{eq:softmax2}
    \alpha_{box}^i=softmax(\frac{Q(t_{box}^{'})\cdot (K(t_{i}^{'}))^T}{\sqrt{d_{k}}}),
\end{equation}
where $Q(t_{cls}^{'})$ and $Q(t_{box}^{'})$ respectively denote queries of class and box tokens. Moreover, $K(t_{i}^{'})$ is the key of the unfolded token sequence, and $d_{k}$ is the dimension of the key and queries.
For each unfolded token, we rank them by their summation of the attention scores and categorize them into a class-related group or a box-related group according to their $\alpha_{cls}^i$ and $\alpha_{box}^i$ values. The class-related group is used for the classification task, and the box-related group is used for the regression task. The details of the task token selection algorithm are shown in Algorithm~\ref{algorithm}.

\subsection{Overall Objective}
The cross-entropy loss ($\mathcal{L}_{CE}$) is used to optimize the classification task, and the smooth $L_1$ loss ($\mathcal{L}_{SmoothL1}$) is used for bounding box regression. 
Moreover, the DN-FPN optimizes the features by jointly using geometric loss (Eq.~\ref{eq:4}) and semantic loss (Eq.~\ref{eq:7}). 
Consequently, the overall loss function is shown in Eq.~\ref{eq:8}, where $\lambda=0.1$ is empirically used to balance the effects of different losses.
\begin{equation}
    \label{eq:8}
    \mathcal{L}_{total}=\mathcal{L}_{CE}+\mathcal{L}_{SmoothL1}+\lambda(\mathcal{L}_{geo}+ \mathcal{L}_{sem}).
\end{equation} 

\begin{algorithm}
    \centering
    \caption{Task Token Selection}\label{algorithm}
    \begin{algorithmic}[1]

        \State \textbf{INPUT :} \hspace{0.1cm} \text{$\alpha_{cls}^i,$} \text{$\alpha_{box}^i,$} \text{$T_{unfold}^{'},$}  \text{$t_{cls}^{'},$} \text{$t_{box}^{'}$}
        \State \text{$G_{c} \gets \{ \}$ : class-related group}
        \State \text{$G_{b} \gets \{ \}$ : box-related group}
        \State \text{$N_u$ : length of the unfolded token sequence }
        \State \textbf{Begin :}
        \State \text{$N_{cls} \gets$  0, } \text{$N_{box} \gets$  0}
        \State \text{$G_{c} \gets$ $t_{cls}^{'}$, } \text{$G_{b} \gets$ $t_{box}^{'}$}
        \State \text{$ Score^i \gets Sum(\alpha_{cls}^i$ , $\alpha_{box}^i)$}
        \State \text{Sort $T_{unfold}^{'}$ according to $Score^i$ in descending order}
        \For {i in range($T_{unfold}^{'},$)}
        \If {$((\alpha_{cls}^i \geq \alpha_{box}^i)$ and $(N_{cls} \leq N_u/2))$ 
        \State or $(N_{box} \geq N_u/2)$ }
            \State $G_{c}\gets t_{i}^{'}$
            \State $N_{cls} \gets N_{cls} + 1$
        \Else
            \State $G_{b} \gets t_{i}^{'}$
            \State $N_{box} \gets N_{box} + 1$
        \EndIf
        
        \EndFor
        \State \textbf{RETURN} \text{$G_{c},$ $G_{b}$}

    \end{algorithmic}
\end{algorithm}

\section{Experiments}
\label{sec:experiment}


\subsection{Experimental Setup}
\subsubsection{Dataset}
To demonstrate the effectiveness of our approach, we use the datasets mainly containing tiny objects, including AI-TOD~\cite{aitod} and VisDrone~\cite{visdrone}. In addition, we evaluate our model on the COCO~\cite{coco} to demonstrate the generalizability of the proposed DNTR. The details of the datasets are introduced as follows. 


\noindent\textbf{AI-TOD.} The AI-TOD dataset includes $28,036$ aerial images and $700,621$ instances. Moreover, $86\%$ of objects in the dataset are smaller than 16 pixels, the largest object is smaller than 64 pixels, and the average object size is only 12.8 pixels. There are $11,214$ images for training, $2,804$ for validation, and $14,018$ for testing. 

\noindent\textbf{VisDrone.} VisDrone consists of $10,209$ drone-shot images and is split into a training set ($6,471$ images), a val set ($548$ images), and a test set ($3,190$ images). There are 10 categories, and the image resolution is $2000 \times 1500$ pixels. We train all the models on the training set and evaluate using the val set since the test set is only available for special challenges (meaning it is not publicly available).


\noindent\textbf{COCO.} This is the most popular dataset in general object detection and contains 80 categories, 118k training images, and 5k validation images. The images in COCO are frontal-view images, unlike those captured from above by drones.

\subsubsection{Implementation Details}
Our work is implemented using the publicly available MMDetection~\cite{mm-det}. During the training phase, we set the input size as $800 \times 800$, $1333 \times 800$, and $1280 \times 800$ for AI-TOD, VisDrone and COCO, respectively. Furthermore, the model is optimized by the SGD optimizer with a learning rate of 0.01, weight decay of 0.0001, and batch size of 2. The number of epochs is set to 36, 12, and 12 for AI-TOD, VisDrone, and COCO, respectively. For hardware, we train and evaluate the models on a single NVIDIA RTX 3090 GPU. For the details of the network architecture, our DNTR model is based on DetectoRS~\cite{detectors} and uses ResNet50 as the backbone. Remarkably, we use Cascade R-CNN instead of HTC~\cite{htc} as the detector of DetectoRS since we do not require the segmentation annotation. 
Additionally, the encoders in DN-FPN that project the features into representations are composed of two convolution layers and two linear layers.

\subsubsection{Evaluation Metrics} 
Following the criterion in~\cite{coco}, the AP, $\text{AP}_{50}$, and $\text{AP}_{75}$ are used to quantitatively evaluate the detection performance on the AI-TOD, VisDrone, and COCO datasets. To further demonstrate the performance in terms of object size, we also use $\text{AP}_{s}$, $\text{AP}_{m}$, and $\text{AP}_{L}$ on COCO. 
In the case of AI-TOD, due to the tiny size of the objects,~\cite{aitod} further subdivides the range of the object into very tiny, tiny, small, and medium. Thus, we use $\text{AP}_{vt}$, $\text{AP}_{t}$, $\text{AP}_{s}$, and $\text{AP}_{m}$ to illustrate the performance.

\begin{table*}
\centering
\normalsize
\caption{Qualitative result on the on the AI-TOD test set. Note that {\color{red}{Red}} refers to the best results, and {\color{blue}{Blue}} refers to the second-best result. * denotes a re-implementation of the results under the same experimental setting.} 
\begin{tabular}{l|c|c|ccccccc} 
\hline
Model                 & Venue     & Backbone  & AP   & $\text{AP}_{50}$ & $\text{AP}_{75}$  & $\text{AP}_{vt}$ & $\text{AP}_{t}$  & $\text{AP}_{s}$  & $\text{AP}_{m}$   \\ 
\hline
\multicolumn{10}{c}{CNN-based models}                                                          \\ 
\hline
FoveaBox~\cite{fcos}                      & TIP2020   & R50      & 8.1  & 19.8 & 5.1  & 0.9  & 5.8  & 13.4 & 15.9  \\
Double Head R-CNN*~\cite{double-head}     & CVPR2020  & R50      & 10.1 & 24.3 & 6.7  & 0.0  & 7.0  & 20.0 & 30.2  \\
Faster R-CNN~\cite{frcnn}                 & TPAMI2017 & R50      & 11.1 & 26.3 & 7.6  & 0.0  & 7.2  & 23.3 & 33.6  \\
YOLOv8s*~\cite{yolov8}                    & Arxiv2023 & R50      & 11.6 & 27.4 & 7.7  & 3.4  & 11.1 & 14.9 & 22.8  \\
QueryDet*~\cite{querydet}                 & CVPR2022  & R50      & 12.2 & 29.3 & 7.9  & 2.4  & 10.5 & 18.5 & 26.3  \\
ATSS~\cite{ATSS}                          & CVPR2021  & R50      & 12.8 & 30.6 & 8.5 & 1.9  & 11.6 & 19.5 & 29.2  \\
Cascade R-CNN~\cite{crcnn}                & CVPR2018  & R50      & 13.8 & 30.8 & 10.5 & 0.0  & 10.5 & 25.5 & 36.6  \\
CenterNetV2*~\cite{centernetv2}           & Arxiv2021 & R50      & 14.1 & 35.7 & 8.5  & 2.9  & 12.7 & 19.7 & 29.5  \\
DetectoRS~\cite{detectors}                & CVPR2021  & R50      & 14.8 & 32.8 & 11.4 & 0.0  & 10.8 & 28.3 & 38.0  \\
YOLOv8l*~\cite{yolov8}                    & Arxiv2023 & R50      & 14.9 & 32.5 & 11.6  & 4.5 & 14.4 & 19.0 & 29.3  \\
DoTD~\cite{dot}                           & CVPRW2021 & R50      & 16.1 & 39.2 & 10.6 & 8.3  & 17.6 & 18.1 & 22.1  \\
Sparse R-CNN~\cite{sparse}                & CVPR2021  & R50      & 16.7 & 38.5 & 11.8 & 8.8  & 17.5 & 18.1 & 19.2  \\
FSANet~\cite{TGRS_FSA}                    & TGRS2022  & R50      & 20.3 & 48.1 & 14.0 & 6.3  & 19.0 & 26.8 & 36.7  \\
HANet~\cite{CSVT_2023}                    & CSVT2023  & R50      & 22.1 & 53.7 & 14.4 & \color{blue}\textbf{10.9} & 22.2 & 27.3 & 36.8  \\
NWD~\cite{nwd}                            & Arxiv2021 & R50      & 20.8 & 49.3 & 14.3 & 6.4  & 19.7 & 29.6 & \color{red}\textbf{38.3}  \\
NWD-RKA~\cite{nwd2}                       & ISPRS2022 & R50      & 23.4 & 53.5 & 16.8 & 8.7  & 23.8 & 28.5 & 36.0  \\
RFLA~\cite{rfla}                          & ECCV2022  & R50   & \color{blue}\textbf{24.8} &  55.2 &  \color{blue}\textbf{18.5} & 9.3 &  \color{blue}\textbf{24.8}  &  \color{blue}\textbf{30.3}  &  \color{blue}\textbf{38.2}   \\ 
\hline
\multicolumn{10}{c}{Transformer-based models}  \\ 
\hline
DETR*~\cite{detr}                      & ECCV2020  & R50      & 2.7  & 10.3 & 0.7  & 0.7  & 2.1  & 3.0  & 12.4  \\
Conditional-DETR*~\cite{cdetr}         & ICCV2021  & R50      & 2.9  & 10.0 & 0.7  & 0.9  & 2.2  & 3.0  & 14.2  \\
DAB-DETR*~\cite{dab-detr}              & ICLR2022  & R50      & 4.9  & 16.0 & 1.7  & 1.7  & 3.6  & 7.0  & 18.0  \\
Deformable-DETR*~\cite{ddetr}          & ICLR2021  & R50      & 17.0 & 45.9 & 8.8  & 7.2  & 17.1 & 22.7 & 28.2  \\
DAB-Deformable-DETR*~\cite{dab-detr}   & ICLR2022  & R50      & 16.5 & 42.6 & 9.9  & 7.9  & 15.2 & 23.8 & 31.9  \\
DINO-Deformable-DETR*~\cite{dino}      & ICLR2023  & R50      & 23.2 & \color{blue}\textbf{56.6} & 15.4 & 9.9  & 23.1 & 29.3 & 37.6  \\ 
\hline
\textbf{DNTR (ours)}                          &  -        & R50      & \color{red}\textbf{26.2} & \color{red}\textbf{56.7} & \color{red}\textbf{20.2} & \color{red}\textbf{12.8}  & \color{red}\textbf{26.4} &\color{red}\textbf{31.0} & 37.0  \\
\hline
\end{tabular}
\label{tab:aitod}
\end{table*}

\subsection{Comparisons with State-of-the-Art Methods}
\subsubsection{Experiment on AI-TOD}
In Table~\ref{tab:aitod}, we compare the performance of DNTR against other strong baselines on the AI-TOD test set. Specifically, these baselines include state-of-the-art methods for remote sensing images~\cite{dot,TGRS_FSA,CSVT_2023,nwd,nwd2,rfla}, as well as the leading detectors for CV tasks~\cite{foveabox,frcnn,detectors,detr,dino}.
Overall, our DNTR outperforms the second-best performance model RFLA~\cite{rfla} by at least 3.5 $\text{AP}_{vt}$ (37.6\%) and 1.6 $\text{AP}_{t}$ (6.5\%). The performance gain is greater when the object size is smaller since this work focuses on detecting tiny objects. To facilitate a detailed analysis of the comparison results, we categorize these baseline methods into CNN-based models and transformer-based models.

Table~\ref{tab:aitod} shows that the performance of all CNN-based detectors is below 11 $\text{AP}_{vt}$, and some of them even get 0 $\text{AP}_{vt}$. 
We attribute the poor performance of the baselines to three reasons:
\begin{enumerate*}[label=(\arabic*)]
\item The instance size is extremely smaller than other datasets, and most of the detection models are evaluated on COCO, which is targeted at generic object detection. Thus, it is expected that the general detection methods perform poorly on AI-TOD. 
\item Due to the redundant noise caused by FPN, the performance of baselines deteriorates on $\text{AP}_{vt}$ due to tiny objects' sensitivity to noise\footnote{The feature map's noise is explored in the ablation study (see Fig.~\ref{fig:psnr_vis}, Fig.~\ref{fig:visual}, and Fig.~\ref{fig:vis_coco}).}.
\item The existing CNN methods are bad at capturing global information. In contrast, with our Trans R-CNN design, the AP can be further improved, especially for the $\text{AP}_{vt}$. 
\end{enumerate*}
Moreover, the existing CNN models may suffer from a trade-off between detecting tiny and large objects. For example, DoTD and Sparse R-CNN are particularly good at handling tiny objects ($\text{AP}_{vt}$) but perform poorly at handling larger ones ($\text{AP}_{m}$). Conversely, ATSS and DetectoRS are good at handling large objects ($\text{AP}_{m}$) but are bad at handling tiny objects ($\text{AP}_{vt}$). Our DNTR outstandingly improves the ability to detect tiny objects while still maintaining performance when detecting large objects.

On the other hand, we also compare the performance of DNTR with the DETR-like models, which are state-of-the-art transformer-based detection models. The transformer-based models perform poorly without the multiscale deformable attention module. Although the benefits from~\cite{ddetr} reap a significant improvement in overall performance, their ability to detect tiny objects still remains below 10 $\text{AP}_{vt}$ as the encoders still extract multiscale features with noise. Furthermore, a DETR-like model's decoder processes box and class information within the same self-attention layer, increasing the difficulty of the detection task. In contrast, the proposed Trans R-CNN ingeniously integrates the box and class tokens by a masking mechanism to boost the performance.

Table~\ref{tab:aitod_large} further investigates the performance with the larger backbone and larger resolution. The larger backbones include ResNet101, ResNeXt~\cite{resnext}, and the Swin transformer~\cite{swin}. The results demonstrate that the AP barely improves for all the larger backbones. For the overall AP, the highest one is DNTR using ResNet101. However, it only gains 0.2 AP, which is a negligible improvement. This is because the larger backbones require more downsampling operations, leading to serious damage to the representation of tiny objects, i.e., they are easily lost after downsampling~\cite{IPG-net, querydet}. Note that the mean size of the objects in AI-TOD is 12.8 pixels, while the size of the largest object in AI-TOD is still smaller than 64 pixels, which is much smaller than other datasets. As such, the larger backbone may be weak at detecting objects of all sizes in AI-TOD.
In line with the observation, DNTR surpasses all other larger models in terms of $\text{AP}_{vt}$. Meanwhile, we increase the images from $800 \times 800$ to $1000 \times 1000$ by bilinear upsampling during training with ResNet50 backbone. According to Table~\ref{tab:aitod_res}, the performance of DetectoRS and DNTR is improved with 3.5 and 2.8 in terms of AP, and 4.2 and 4.5 in terms of $\text{AP}_{vt}$, respectively. This suggests that increasing the input resolution brings more advantages to tiny objects than using a larger backbone. 

\begin{table}
\caption{A comparison of the DNTR with a larger backbone on the AI-TOD test set.} 
\centering
\begin{tabular}{l|c|ccccc}
\hline
Model & Backbone & AP  & $\text{AP}_{50}$ & $\text{AP}_{vt}$ & $\text{AP}_{t}$  & $\text{AP}_{s}$ \\
\hline
DNTR & Swin-L~\cite{swin}     & 18.3 & 37.8 & 7.2 & 14.8 & \textbf{32.6}  \\
DNTR & R-NeXt~\cite{resnext}  & 23.5 & 50.9 & 10.0  & 22.9 & 28.5 \\
DNTR & R101     & \textbf{26.4} & 56.3  & 11.8  & 26.3 & 31.4  \\
DNTR & R50      & 26.2  & \textbf{56.7} & \textbf{12.8} & \textbf{26.4} & 31.0  \\
\hline
\end{tabular}
\label{tab:aitod_large}
\end{table}

\begin{table}
\caption{A comparison of DNTR with a larger resolution on the AI-TOD test set. Note that L denotes the resolution with $1000\times1000$.}
\centering
\begin{tabular}{l|cccccc}
\hline
Model         &  AP &  $\text{AP}_{50}$ &  $\text{AP}_{vt}$ & $\text{AP}_{t}$  & $\text{AP}_{s}$  & $\text{AP}_{m}$\\
\hline
DetectoRS~~~   & 14.8  & 32.8  & 0.0  & 10.8  & 28.3 & 38.0 \\
DetectoRS-L       & 18.3  & 44.9  & 4.2  & 16.9  & 25.2 & 37.8 \\
\textbf{Improvements}  &\textbf{+3.5} &\textbf{+12.1}  &\textbf{+4.2} &\textbf{+6.1} &\textbf{-3.1} &\textbf{-0.2} \\ 
\hline
DNTR     & 26.2   & 56.7    & 12.8  & 26.4  & 31.0  & 37.0 \\
\textbf{DNTR-L}  & \textbf{29.0} & \textbf{58.3}  & \textbf{17.3} & \textbf{29.2} & \textbf{33.5} & \textbf{40.5} \\
\textbf{Improvements}  &\textbf{+2.8} &\textbf{+1.6}&\textbf{+4.5} 
&\textbf{+2.8} &\textbf{+2.5} &\textbf{+2.5}\\
\hline
\end{tabular}
\label{tab:aitod_res}
\end{table}


\subsubsection{Experiments on VisDrone} 
Table~\ref{tab:vis} shows the performance of different models on the VisDrone val set. The evaluation tool and the dataset settings follow the official guidelines~\cite{vis-tool}.
Since drone-shot images require cropping strategies for making foreground features become more noticeable after resizing the sub-images, we conduct an experiment with and without cropping strategies. The top part of Table~\ref{tab:vis} shows that DNTR reaches state-of-the-art performance without using any cropping strategies. DNTR without cropping strategies respectively outperforms the DetectoRS (the degenerated model of DNTR) by 10.8 AP (48.4\%) and 13.7 $\text{AP}_{50}$ (34.2\%). 
However, with the cropping strategies (bottom part of Table~\ref{tab:vis}), our DNTR's performance slightly lags behind that of the state-of-the-art method, UFPMP~\cite{UFPMP}, due to its utilization of a two-stage pipeline instead of an end-to-end model.  
Specifically, they use one network for mosaic packaging (pre-possessing) and the other network for object detection (post-processing).
Thus, we also apply our DN-FPN modules on UFPMP~\cite{UFPMP} to utilize their cropping strategy and achieve state-of-the-art performance. The results manifest that our DNTR model (or only with the DN-FPN module) is capable of highlighting tiny objects from images.

\begin{table}[ht]
\caption{Quantitative result on the VisDrone val set. Note that {\color{red}{Red}} refers to the best results, and {\color{blue}{Blue}} refers to the second-best result. * denotes a re-implementation of the results under the same experimental setting. CS and MP indicate the cropping strategies and mosaic packing, respectively.}
\centering
\small
\begin{tabular}{l|ccc|c} 
\hline
Model                  & AP    & $\text{AP}_{50}$  & $\text{AP}_{75}$  & CS  \\ 
\hline
Faster R-CNN~\cite{frcnn}           & 21.4  & 40.7  & 19.9  & -                \\
Cascade R-CNN~\cite{crcnn}           & 22.6  & 38.8  & 23.2  & -                \\
DetectoRS*~\cite{detectors}               & 22.3  & 40.1  & 22.3  & -   \\
Yolov5~\cite{yolov5}              & 24.1  & 44.1  & 24.7  & -\\
CEASC~\cite{CEASC}                 & 28.7  & 50.7  & 28.4  & -                \\
SDP~\cite{TGRS_SDP}               & \color{blue}\textbf{30.2}  & \color{blue}\textbf{52.5}  & \color{blue}\textbf{30.6}  & -   \\
\textbf{DNTR}                & \color{red}\textbf{33.1}  & \color{red}\textbf{53.8} & \color{red}\textbf{34.8} & - \\
\hline
RetinaNet~\cite{retinanet}              & 26.2  & 44.9  & 27.1  & Evenly                \\
ClusDet~\cite{clusterdet}                & 26.7  & 50.6  & 24.7  & Cluster          \\
GLSAN~\cite{GLSAN}                  & 32.5  & 55.8  & 33.0  & Density          \\
CZDet~\cite{density-crop}        & 33.2  & 58.3  & 33.2  & Density         \\
QueryDet~\cite{querydet}               & 28.3  & 48.1  & 28.8  & Evenly                \\
AMRNet~\cite{amrnet}                 & 31.7  & 52.7  & 33.1  & Adaptive         \\
DetectoRS*~\cite{detectors}     & 29.4  & 49.3 & 30.2 & MP  \\ 
DNTR     & 34.4  & 57.9 & 35.3 & MP \\
UFPMP~\cite{UFPMP}              & \color{blue}\textbf{36.6}  & \color{blue}\textbf{62.4}  & \color{blue}\textbf{36.7}  & MP  \\
\textbf{UFPMP w/ DN-FPN}        & \color{red}\textbf{37.8}  & \color{red}\textbf{62.7} & \color{red}\textbf{38.6}  & MP     \\
\hline
\end{tabular}
\label{tab:vis}
\end{table}

\subsubsection{Experiments on COCO} 
To further demonstrate the generalization ability of DNTR, we also evaluate the proposed method on the COCO dataset, as shown in Table~\ref{tab:coco}. Based on the statistics of dataset~\cite{aitod, coco, sm}, the absolute size and relative size of most instances in COCO are far greater than those of the other tiny object datasets (e.g., AI-TOD). For a fair comparison, the resolution of DNTR we used in COCO is $1280 \times 800$ with ResNet50 as the backbone, which is the same as all the counterparts in Table~\ref{tab:coco}. Remarkably, the $\text{AP}_{L}$ of DNTR still has a competitive advantage over CNN-based and DETR-like models. The results manifest that the proposed method is feasible and effective for objects of a general size.


\begin{table*}
\caption{Quantitative result on the COCO val set. Note that {\color{red}{Red}} refers to the best results, and {\color{blue}{Blue}} refers to the second-best result. * denotes a re-implementation of the results under the same experimental setting.}
\centering
\normalsize
\begin{tabular}{l|c|c|c|cccccc}
\hline  
Method  & Venue  & Backbone & Epochs & AP   & $\text{AP}_{50}$ & $\text{AP}_{75}$ & $\text{AP}_{S}$  & $\text{AP}_{M}$  & $\text{AP}_{L}$   \\
\hline
Faster R-CNN~\cite{frcnn}  & TPAMI2017 & R50  & 109    & 42.0 & 62.1 & 45.5 & 26.6 & 45.5 & 53.4 \\
DETR-DC5~\cite{detr}  & ECCV2020   & R50    & 500    & 43.3 & 63.1 & 45.9 & 22.5 & 47.3 & 61.1 \\
Conditional-DETR-DC5~\cite{cdetr} & ICCV2021  & R50    & 50     & 43.8 & 64.4 & 46.7 & 24.0   & 47.6 & 60.7 \\
DAB-DETR-DC5~\cite{dab-detr}  & ICLR2022  & R50 & 50     & 44.5 & 65.1 & 47.7 & 25.3 & 48.2 & 62.3 \\
Deformable-DETR~\cite{ddetr}   & ICCV2021 & R50     & 50     & 46.2 & 65.2 & 50.0 & \color{blue}\textbf{28.8} & 49.2 & 61.7 \\
DetectoRS*~\cite{detectors}  & CVPR2021 & R50  & \textbf{12}     & 46.1 & 64.6 &\color{blue}\textbf{50.1} & 28.0 &\color{blue}\textbf{50.1} & 59.4 \\
DN-DETR-DC5~\cite{dndetr} & CVPR2022  & R50   & 50     & \color{blue}\textbf{46.3} & \color{blue}\textbf{66.4} & 49.7 & 26.7 & 50.0 & \color{blue}\textbf{64.3} \\
\hline
\textbf{DNTR} & - & R50   & \textbf{12}     & \color{red}\textbf{47.9} & \color{red}\textbf{67.0} & \color{red}\textbf{50.7} & \color{red}\textbf{30.3} & \color{red}\textbf{50.8} & \color{red}\textbf{64.5} \\
\hline
\end{tabular}
\label{tab:coco}
\end{table*}

\subsection{Ablation Studies}
\subsubsection{Effectiveness of Individual Component}
Table~\ref{tab:abla} shows the ablation studies of different components in DNTR on the AI-TOD test set, including DN-FPN, Mask Transformer Encoder (MTE), and Task Token Selection (TTS). The results demonstrate that DN-FPN context information propagation within FPN boosts the performance of tiny object detection. 
Compared to the baseline, which is NWD~\cite{nwd}, the performance is improved by 4.8 AP (23.0\%), 4.0 $\text{AP}_{vt}$ (62.5\%), 6.3 $\text{AP}_{t}$ (32.0\%), and 0.3 $\text{AP}_{s}$ (1.0\%) with the DN-FPN only. Moreover, MTE increases $\text{AP}_{vt}$ from 10.4 to 12.7 by better leveraging the global information. 
Lastly, TTS can raise the AP slightly and reach state-of-the-art performance. For the model complexity, DNTR improves the overall performance considerably with a bit of FLOPs overhead.

\begin{table}[htbp]
\caption{The ablation study of our architecture on the AI-TOD test set. Please note that the MTE indicates the Mask Transformer Encoder, and TTS indicates the Task Token Selection.}
\centering
\small
\begin{tabular}{c|c|c|ccc|c} 
\hline
DN-FPN      & MTE        & TTS        & AP      & $\text{AP}_{vt}$     & $\text{AP}_{t}$      & FLOPs (G)\\ 
\hline
            &            &            & 20.8  & 6.4       & 19.7      & 159.7\\
\checkmark  &            &            & 25.6  & 10.4      & 26.0               & 159.7\\
 \checkmark & \checkmark &            & 25.9  & 12.7      & 26.0                & 164.0 \\
& \checkmark & \checkmark & 25.7  & 12.5  & 25.9   & 167.5\\
\checkmark  & \checkmark & \checkmark & \textbf{26.2}   & \textbf{12.8} & \textbf{26.4}  & 167.5\\
            
\hline
\end{tabular}
\\
\label{tab:abla}
\end{table}

\subsubsection{Complexity Analysis of the DN-FPN module}
Table~\ref{tab:fpn} compares the proposed DN-FPN with other methods that enhance FPN modules in terms of AP, model size, and FLOPs on the AI-TOD dataset. The results demonstrate that our DN-FPN does not use any extra parameters and FLOPs on FPN while improving AP and $\text{AP}_{vt}$ by 10.7 (71.8\%) and 7.9 (316.0\%) as compared to the FA\textsuperscript{2}M module. This is because the geometric and semantic encoders in DN-FPN are only used in the training stage for producing denoising features and are discarded during the inference. In contrast, to enhance the spatial information of tiny objects,~\cite{TGRS_FSA} aggregates the FPN features by an extra offset network, which learns the alignment between the high-level and low-level features.

\begin{table}
\centering
\small
\caption{Comparisons of the computational complexity of DN-FPN with other enhanced FPN methods on the AI-TOD test set. Note that the parameters and FLOPs results only show the additional module based on FPN (see Table~\ref{tab:abla} for total FLOPs). }

\begin{tabular}{l|ccc|cc}
\hline
Method                 & AP   & AP50  & $\text{AP}_{vt}$ & \begin{tabular}[c]{@{}c@{}}Param.\\(M)\end{tabular} & \begin{tabular}[c]{@{}c@{}}FLOPs\\(G)\end{tabular}  \\ 
\hline
FaPN~\cite{FaPN}    & 14.8 & 25.9   & -    & 6.4    & 25.3 \\
FA\textsuperscript{2}M ~\cite{TGRS_FSA} & 14.9 & 26.2   & 2.5  & 2.7    & 10.1 \\
DN-FPN           & \textbf{25.6} & \textbf{56.1} & \textbf{10.4} & \textbf{0.0}  & \textbf{0.0}   \\
\hline
\end{tabular}
\label{tab:fpn}
\end{table}



\subsubsection{Effectiveness of the DN-FPN module}
Since the proposed DN-FPN improves FPN without introducing additional FLOPs and parameters, FPN in other models can be easily augmented by DN-FPN to enhance performance. Table~\ref{tab:abla-2} shows the performance of different detection methods with the proposed DN-FPN. The results show that the proposed DN-FPN benefits different models, including one-stage models (i.e., FoveaBox, ATSS, YOLOv8s) and two-stage models (Faster R-CNN, Double Head R-CNN, DetectoRS). As such, DN-FPN is a better neck and can be used as a plug-and-play module. It is worth noting that DetectoRS has a recursive feature pyramid structure (two FPNs). Thus, it can benefit from the noise reduction capability of DN-FPN twice, improving 10.8 AP and the 10.4 $\text{AP}_{vt}$. 

\begin{table}
\centering
\caption{Comparison between different models with the proposed DN-FPN on the AI-TOD test set.}
\small
\begin{tabular}{l|cccc}
\hline
Model                        & AP    & $\text{AP}_{vt}$  & $\text{AP}_{t}$   & $\text{AP}_{s}$     \\
\hline
FR-CNN~\cite{frcnn}                      & 11.1   & 0.0    & 7.2  & 23.3 \\
FR-CNN w/ DN-FPN            & 12.6    & 1.7    & 9.1  & 25.6  \\
\textbf{Improvements}  &{\textbf{+1.5}} &{\textbf{+1.7}} &{\textbf{+1.9}} &{\textbf{+2.3}} \\
\hline
FoveaBox~\cite{foveabox}                          & 8.1    & 0.9  & 5.8  & 13.4  \\
FoveaBox w/ DN-FPN                & 13.9  & 1.8  & 11.1 & 21.0 \\
\textbf{Improvements}  &\textbf{+5.8} &\textbf{+0.7} &\textbf{+5.3} &\textbf{+7.6}  \\
\hline
ATSS~\cite{ATSS}                              & 12.8   & 1.9  & 11.6 & 19.5  \\
ATSS w/ DN-FPN                    & 17.9 & 3.7  & 16.4 & 25.3  \\
\textbf{Improvements}  &\textbf{+5.1} &\textbf{+1.8} &\textbf{+4.8} &\textbf{+5.8}  \\
\hline
DH R-CNN~\cite{double-head}      & 10.1   & 0    & 7.0  & 20.0  \\
DH R-CNN w/ DN-FPN       & 13.0  & 2.8    & 7.2  & 28.4  \\
\textbf{Improvements}  &\textbf{+2.9} &\textbf{+2.8} &\textbf{+0.2} &\textbf{+8.4} \\
\hline
YOLOv8s~\cite{yolov8}      & 11.6  & 3.4  & 11.1  & 14.9  \\
YOLOv8s w/ DN-FPN          & 18.7  & 6.7  & 15.8  & 24.7  \\
\textbf{Improvements}  &\textbf{+4.9} &\textbf{+3.3} &\textbf{+4.7} &\textbf{+9.8} \\
\hline
DetectoRS~\cite{detectors}     & 14.8  & 0.0    & 10.8 & 28.3 \\
\textbf{DetectoRS w/ DN-FPN}    & \textbf{25.6}  & \textbf{10.4} &\textbf{26.0}& \textbf{29.9} \\
\textbf{Improvements}  &\textbf{+10.8} &\textbf{+10.4} &\textbf{+15.2} &\textbf{+1.6}\\ 
\hline
\end{tabular}
\label{tab:abla-2}
\end{table}

\subsubsection{Efficiency Analysis of the DNTR}
In evaluating the effectiveness of different object detection models, we present a comparative analysis in terms of inference speed (measured in frames per second, FPS) and computational complexity (quantified as FLOPs) between our proposed DNTR and other established baseline methods, as shown in Table~\ref{tab:fps}. Our DNTR demonstrates superior AP across all tested models. However, it does not achieve the highest inference speed. Notably, models based on the YOLO and DETR architectures exhibit higher FPS rates but compromise on AP scores, especially in the context of $\text{AP}_{vt}$, which measures the precision for detecting very tiny objects.

It is important to emphasize that the primary goal of our research is to enhance the detection performance of tiny objects through the DNTR model. Considering the balance between accuracy and inference speed, we further integrate DN-FPN with YOLOv8s by substituting the PAN-FPN~\cite{panet} with our DN-FPN. The integration results, as detailed in Table~\ref{tab:fps}, reveal that YOLOv8s equipped with DN-FPN not only matches the original speed of YOLOv8s but also significantly improves detection accuracy, with increases of 61.2\%, 59.1\%, and 97.1\% in AP, $\text{AP}_{50}$, and $\text{AP}_{vt}$ scores, respectively. In conclusion, the proposed DNTR sets a new benchmark for tiny object detection in terms of accuracy, while a modified YOLOv8s integrated with DN-FPN offers a competitive solution that does not compromise on speed.

\begin{table}[h]
\centering
\caption{Comparison of the Efficiency of DNTR on the AI-TOD test set. Note that DN-FPN is eliminated during the inference stage, thus incurring no additional cost in terms of FPS and FLOPs.}
\begin{tabular}{l|ccc|cc} 
\hline
Method   & AP   & $\text{AP}_{50}$  & $\text{AP}_{vt}$     & FPS    &  FLOPs(G) \\ 
\hline
YOLOv8s~\cite{yolov8}        & 11.6    & 27.4    & 3.4     & \textbf{169.5}  &  \textbf{28.8} \\
YOLOv8l~\cite{yolov8}        & 14.9    & 32.5    & 4.5    & 61.72  & 165.7 \\
Deformable-DETR~\cite{ddetr}           & 17.0    & 45.9    & 7.2     & 15.0   & 196.0   \\
NWD-RKA~\cite{rfla}           & 23.4    & 53.5    & 8.7     & 13.5   & 159.7    \\
\hline
YOLOv8s w/ DN-FPN     & 18.7    & 43.6    & 6.7    & \textbf{169.5}  &  \textbf{28.8}   \\
DNTR   & \textbf{26.2} & \textbf{56.7} &\textbf{12.8} &  11.9 & 167.5   \\
\hline
\end{tabular}
\label{tab:fps}
\end{table}


\subsubsection{Coefficient of the Training Objective} 
During training, we observed that the summation of the geometric and semantic loss tends to dominate the overall loss, which neglects the classification and detection loss, resulting in a reduction of the AP. Hence, we employ a coefficient $\lambda$ to restrict the influence of geometric and semantic loss, as shown in Table~\ref{tab:coef}. Accordingly, an elevation of $\lambda$ leads to a decrease in AP, yet the adjustment of the coefficient $\lambda$ exhibits low sensitivity to overall performance. Consequently, we select $\lambda$ to 0.1, which achieves the highest $\text{AP}_{vt}$.

\begin{table}
\centering
\small
\caption{Ablation on the coefficient of the training loss.}
\begin{tabular}{c|ccccccc} 
\hline
$\lambda$  & AP   & $\text{AP}_{50}$  & $\text{AP}_{75}$  & $\text{AP}_{vt}$  & $\text{AP}_{t}$   & $\text{AP}_{s}$   & $\text{AP}_{m}$    \\ 
\hline
0.05 & 25.4 & 54.3 & 19.5 & 11.4 & 25.3 & 30.2 & 36.5  \\
0.07 & 25.8 & 55.8 & \textbf{20.3} & 12.7 & 25.7 & 29.7 & \textbf{37.3} \\
\textbf{0.1} & \textbf{26.2} & \textbf{56.7} & 20.2 & \textbf{12.8} & \textbf{26.4} & \textbf{31.0} & 37.0  \\
0.5 & 25.5 & 55.5 & 19.9 & 12.7 & 25.1 & 29.5 & 36.4  \\
0.7 & 25.3 & 55.3 & 18.9 & 11.9 & 25.0 & 28.9 & 36.3  \\
1   & 24.9 & 54.2 & 19.3 & 10.8 & 25.2 & 27.7 & 34.6  \\
\hline
\end{tabular}
\label{tab:coef}
\end{table}

\subsubsection{Unfolding Order and the Number of Unfolded Tokens}
Here, we investigate the effect of shuffle unfolding and the number of unfolded tokens. Table~\ref{tab:shuffle} compares the raster scan order and the proposed shuffle order, which manifests that these two approaches obtain a similar AP under the basic token number (1x), i.e., 16 tokens within an RoI feature. However, when the oversampling is applied on tokens, the AP increases more with the proposed shuffle unfolding. This is because the shuffle unfolding operation brings more variety to object representation.
Consequently, we use shuffle order with 64 tokens as our detail setting in the Trans R-CNN detector.
\begin{table}[h]
\centering
\caption{The unfolding order and the number of unfolded tokens on the AI-TOD test set. Note that the 1x and 4x denote the oversampling rate of tokens ($r$).}
\small
\begin{tabular}{l|c|cccc} 
\hline
\begin{tabular}[c]{@{}l@{}}Token number\end{tabular} & Order & AP  & $\text{AP}_{vt}$   &$\text{AP}_{t}$       \\ 
\hline
16 (1x)  & Raster scan  & 24.6    & 11.3  & 25.1  \\
16 (1x)  & Shuffle      & 24.8    & 9.8  & 25.3\\
64 (4x)   & Shuffle   & \textbf{26.2} &\textbf{12.8}   & \textbf{26.4} \\
\hline
\end{tabular}
\label{tab:shuffle}
\end{table}

\subsection{Visualization}  
To further validate the denoising capability of our DN-FPN, we propose to leverage the Peak Signal-to-Noise Ratio (PSNR) to evaluate the noise level of the feature maps. By first creating a target image (zero matrix) with the shape $H \times W \times 1$, where $H$ and $W$ denote height and width, respectively, we project the center points of the ground truth bounding boxes onto the target image. The Gaussian kernel (with a standard deviation parameter of 4) is applied to each center point on the target image, transforming it into the target feature.

Next, we select the bottom feature map, characterized by a dimension of $H \times W \times 256$, from the top-down path of the FPN as the distorted image used for PSNR calculation, leveraging its heightened utility in tiny object detection. To compute the PSNR with target features, we select the maximum value among 256 channels at each position to transform the feature maps to $H \times W \times 1$. By doing so, the feature map is transformed into the \textbf{response feature}.
Consequently, we compute the PSNR between the target feature and the response feature. A higher PSNR reflects a higher feature quality, resulting in better detection ability for tiny objects.

\begin{figure}[h] 
\centering 
        \includegraphics[width=1.0\columnwidth]{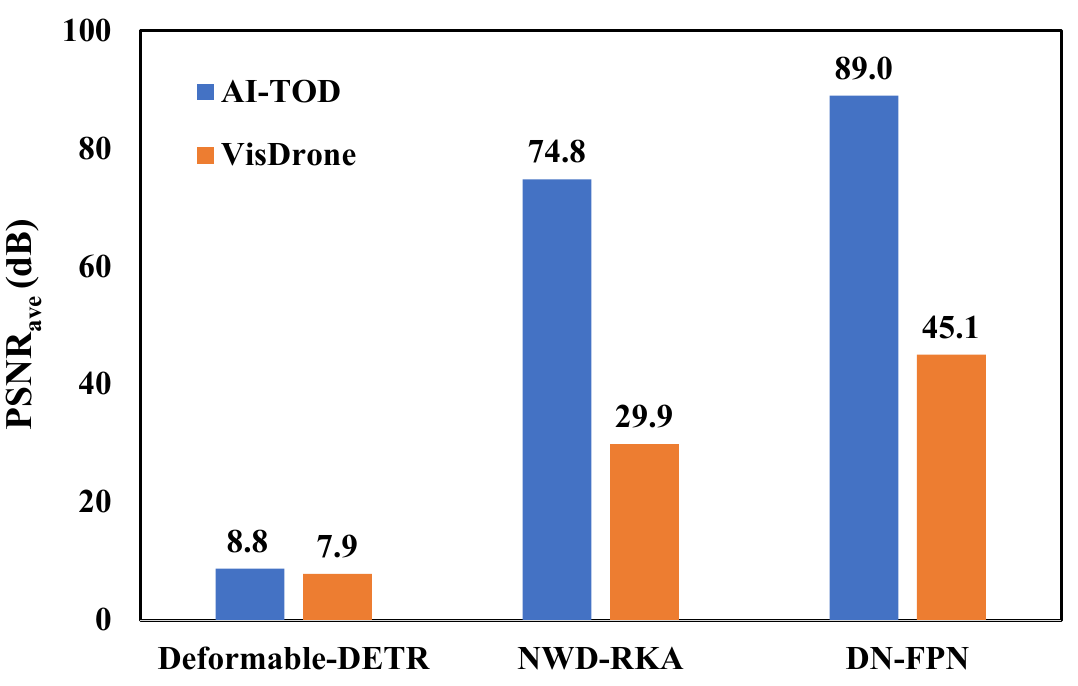}
        \caption{Noise analysis of feature maps between different methods on the AI-TOD test set and VisDrone val set.}
        \label{fig:psnr_vis}
\end{figure}

\begin{figure*}[t]
\centering
    \includegraphics[width=1.5\columnwidth]{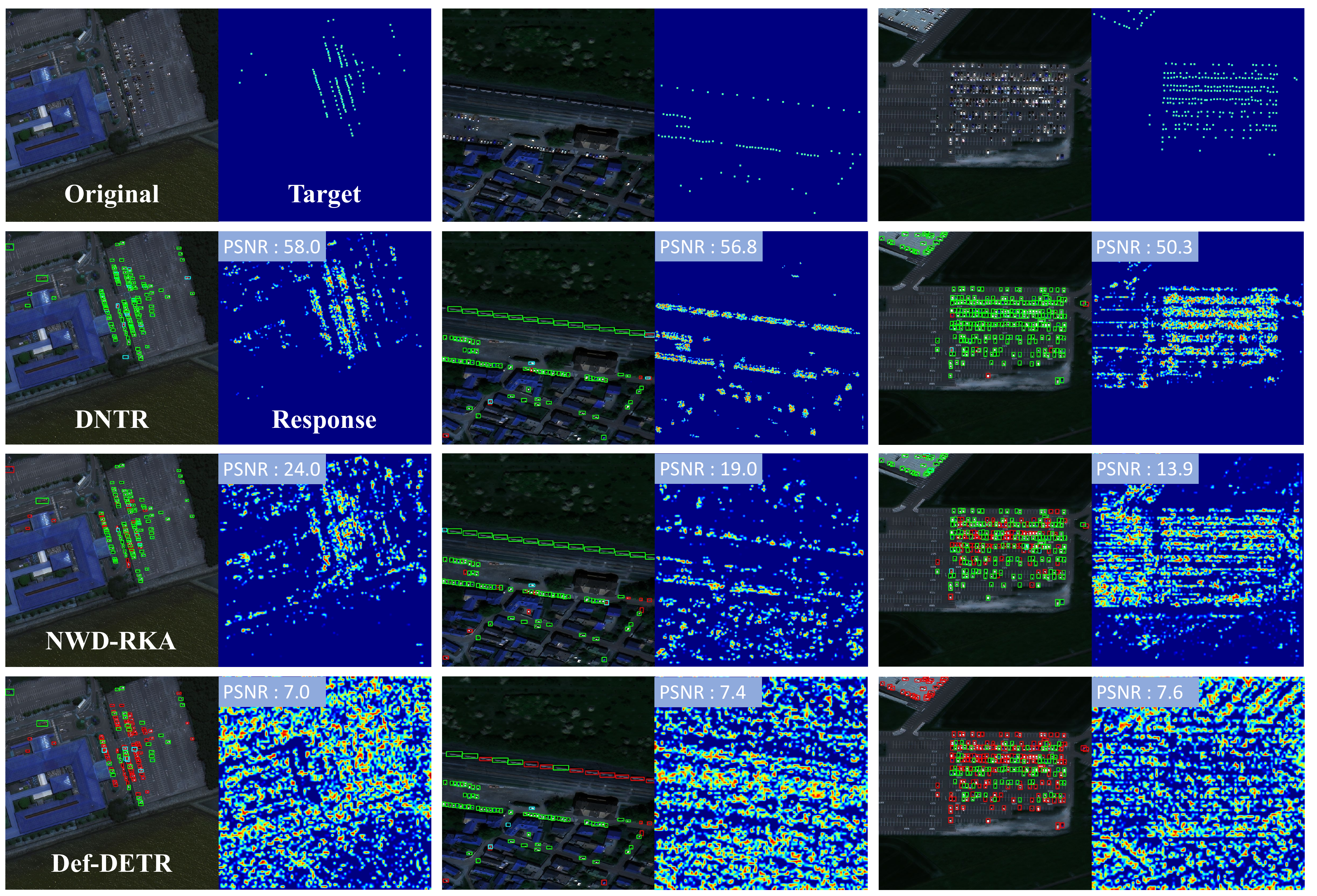}
    \caption{Visualization of model results on the AI-TOD test set. The first column shows the original images and their target features. The rest of the columns are within detection results and corresponding response features. We compare the results of Deformable-DETR, NWD-RKA, and DNTR. The green, cyan, and red boxes indicate true positives (TP), false positives (FP), and false negatives (FN), respectively, while Def-DETR stands for Deformable-DETR. Our DNTR evidently outperforms in both detection results and the quality of response features.}
    \label{fig:visual}
\end{figure*}

The average PSNR of the dataset is represented by $PSNR_{ave}$. Fig.~\ref{fig:psnr_vis} compares $PSNR_{ave}$ of our DNTR with different methods. Our approach significantly outperforms Deformable-DETR~\cite{ddetr} and NWD-RKA~\cite{nwd2} on both AI-TOD and VisDrone datasets. The performance of Deformable-DETR is much worse than that of other methods, which aligns with our earlier description. 
Consequently, we visualize the target and response features of Deformable-DETR, NWD-RKA, and DNTR on the AI-TOD dataset in Fig.~\ref{fig:visual}. 
The results demonstrate that the proposed DN-FPN reduces noise effectively, while the proposed Trans R-CNN concentrates more on foreground objects than the background. In contrast, the response features of Deformable-DETR and NWD-RKA show attention to the background, which is regarded as the noise in the feature maps. Moreover, DNTR better detects tiny bounding boxes than its counterparts since the results yield more truth positives and fewer false negatives. Another observation is that Deformable-DETR contains a lot of background responses since DETR-like methods do not apply the RPN to filter out the unimportant areas, leading to false positives~\cite{DQ-detr}.

\begin{figure}[t] 
\centering 
                \includegraphics[width=\columnwidth]{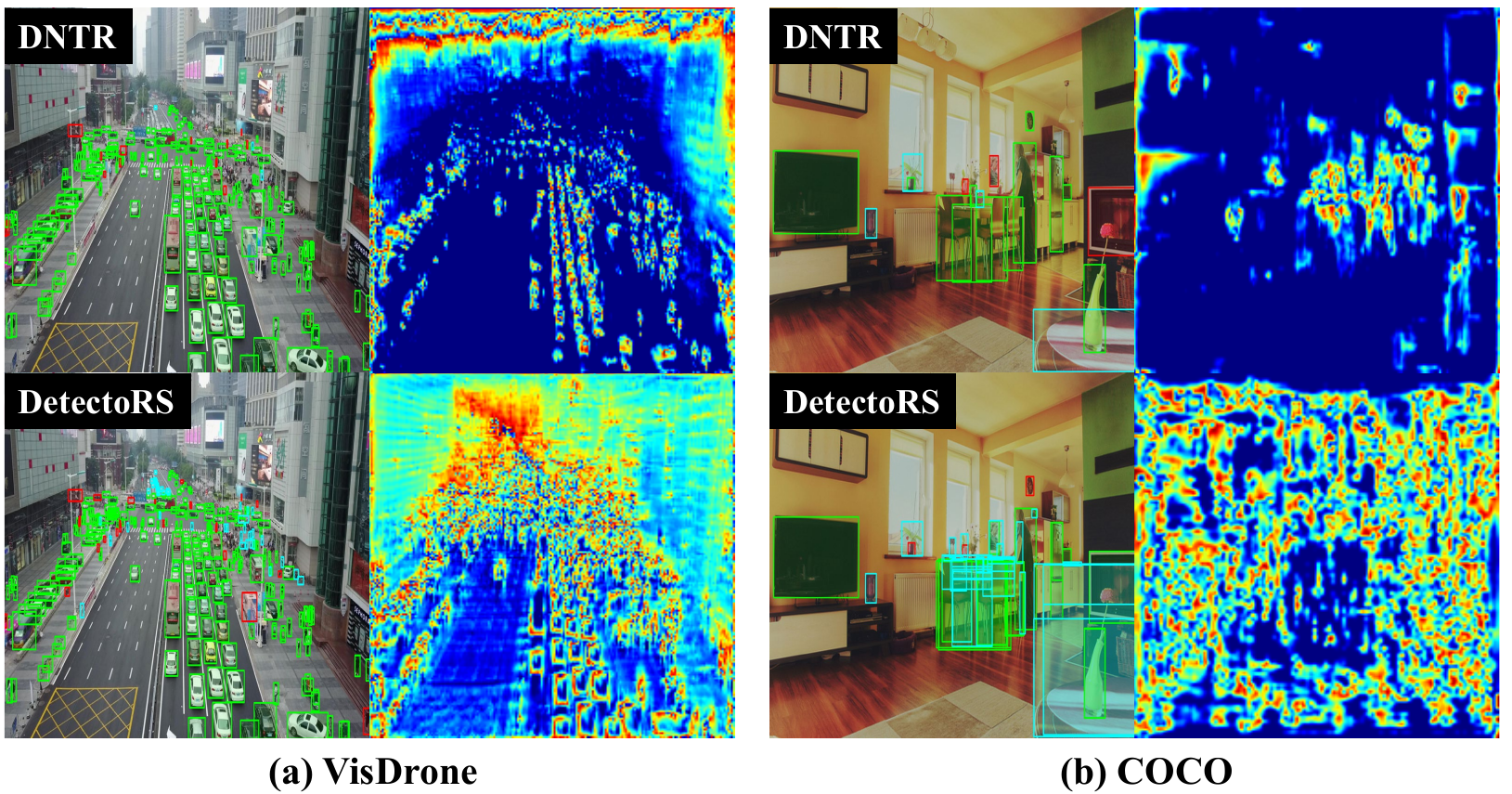}
                \caption{Visualization of the model results on the (a) VisDrone val set and (b) COCO val set. The upper and lower rows display the detection and feature response results of DNTR and the DetectoRS, respectively. Note that the green, red, and cyan boxes denote the TP, FP, and FN of the detection results.}
                \label{fig:vis_coco}
\end{figure}

To demonstrate DNTR's robustness, we also visualize the detection and their corresponding response feature results on the VisDrone and COCO datasets, as illustrated in Fig.~\ref{fig:vis_coco}(a) and Fig.~\ref{fig:vis_coco}(b), respectively. Under the findings from the detection results, our DNTR consistently obtains fewer false positives and false negatives in the detection results across both the VisDrone val set and COCO val set. Moreover, the response features of DetectoRS still exhibit a significant response to the cluttered background. In contrast, our model can effectively disregard the background and focus solely on the relevant foreground information. This observation is consistent with the findings obtained from the AI-TOD dataset.

\section{Conclusion and future works}
\label{sec:conc}
We introduce DNTR, a two-stage model optimized for tiny object detection. The design of the DN-FPN module adeptly mitigates noise through geometric-semantic contrastive learning. Simultaneously, the Trans R-CNN facet excels in global information aggregation via the mask transformer encoder and finetunes discriminative feature extraction through the innovative task token selection mechanism. Experimental results show that the proposed DNTR outstripped contemporaries on the AI-TOD and VisDrone datasets, recording AP scores of 26.2\% and 33.1\%, respectively, thereby asserting its state-of-the-art stature. While our development signals a significant stride in tiny object detection, it also casts a spotlight on the prevailing challenges faced by full transformer detectors in this specialized domain, as evidenced by the relative underperformance of DETR-like techniques on the AI-TOD dataset. We recognize this as a critical frontier for further advancements and pledge to deepen our exploration in this direction, harboring aspirations to enhance DNTR further. In summary, DNTR not only carves out a promising trajectory for future research but establishes a benchmark in tiny object detection, nudging the scientific community towards innovative solutions that bridge existing gaps in the field.

\section*{Acknowledgement}
This work is partially supported by the National Science and Technology Council, Taiwan under Grants, NSTC-112-2628-E-002-033-MY4, NSTC-112-2634-F-002-002-MBK, NSTC-112-2221-E-A49-059-MY3, and NSTC-112-2221-E-A49-094-MY3.



{\small
\bibliographystyle{IEEEtran}
\bibliography{egbib}}

\vspace{-4mm}
\begin{IEEEbiography}[{\includegraphics[width=1in,height=1.25in, clip,keepaspectratio]{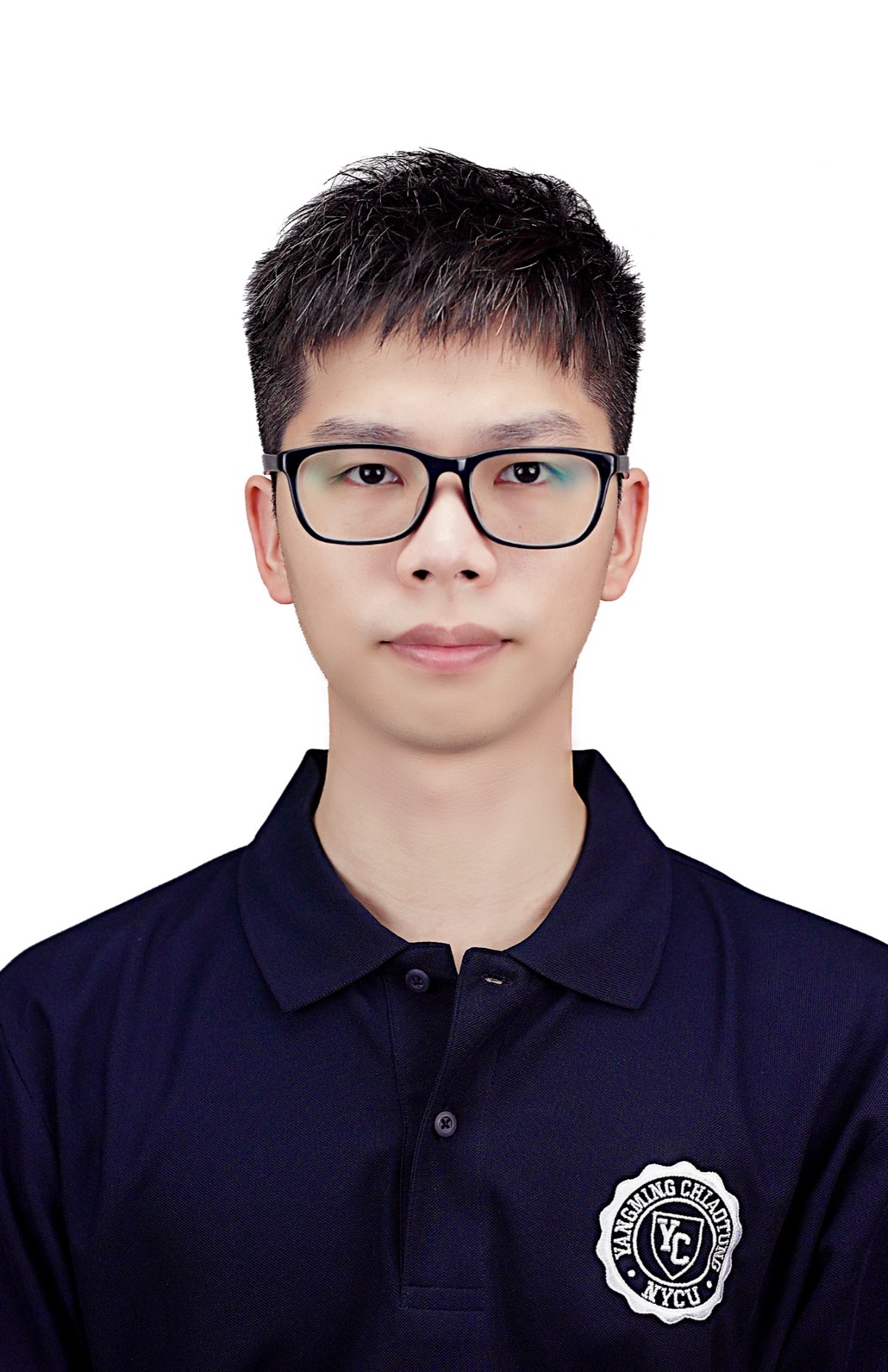}}]{Hou-I Liu} received the B.S. degree from the Department of Materials Science and Engineering, National Tsing Hua University (NTHU), Hsinchu, Taiwan, R.O.C., in 2016, the M.S. degree in electrical engineering from National Taipei University of Technology (NTUT) in 2018. He is currently a Ph.D. candidate at National Yang Ming Chiao Tung University (NYCU). His research interests include computer vision, object detection on remote sensing, and high-resolution images.
\end{IEEEbiography}

\vspace{-4mm}
\begin{IEEEbiography}
[{\includegraphics[width=1in,height=1.25in,clip,keepaspectratio]{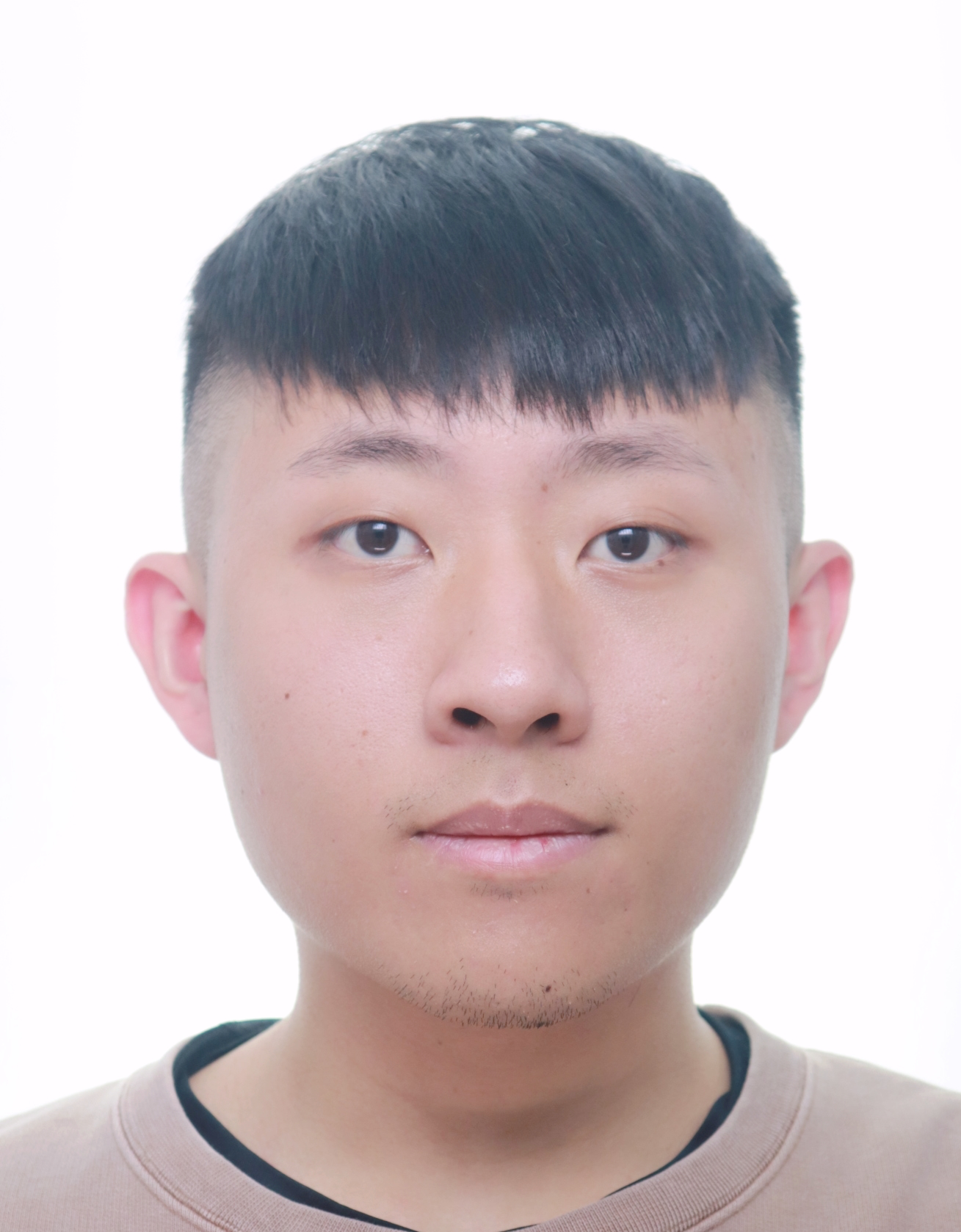}}]{Yu-Wen Tseng} received the B.S. degree from the Undergraduate Honors Program of Electrical Engineering and Computer Science, National Yang Ming Chiao Tung University (NYCU), Hsinchu, Taiwan, R.O.C., in 2021. He is now a master's student at NYCU. His research interests include visual perception in 2D and 3D scenarios.
\end{IEEEbiography}

\vspace{-3mm}
\begin{IEEEbiography}
[{\includegraphics[width=1in,height=1.25in,clip,keepaspectratio]{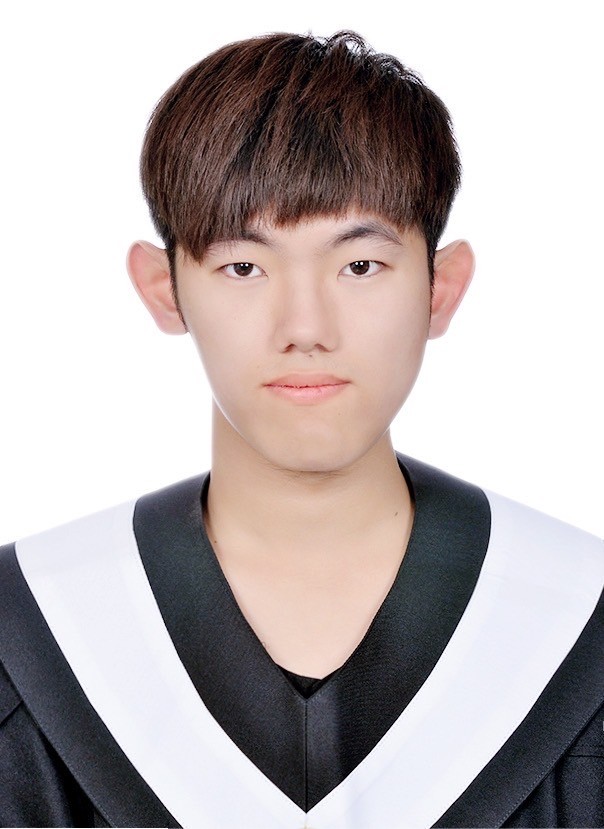}}]{Kai-Cheng Chang} received the B.S. degree from the Department of Electrical Engineering, National Chung Hsing University (NCHU), Taichung, Taiwan, R.O.C., in 2021. He is now a master's student at National Yang Ming Chiao Tung University (NYCU). His research interests include domain adaptation and traffic light detection.
\end{IEEEbiography}

\begin{IEEEbiography}
[{\includegraphics[width=1in,height=1.25in,clip,keepaspectratio]{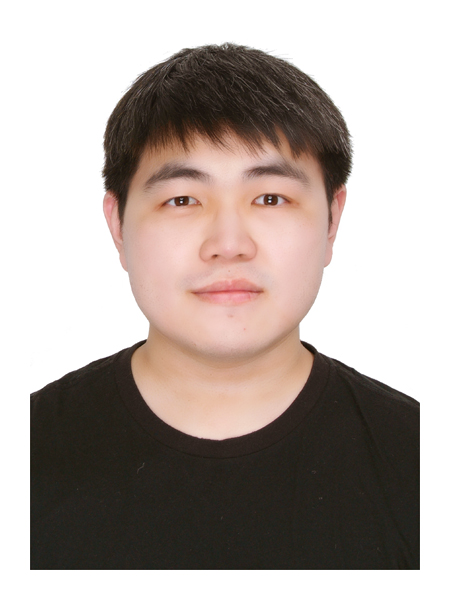}}]{Pin-Jiun Wang} received the B.S. degree from the Department of Electrical Engineering, National Yang Ming Chiao Tung University (NYCU), Hsinchu, Taiwan, R.O.C., in 2024. 

His research interests include deep learning and crowded object detection.
\end{IEEEbiography}

\begin{IEEEbiography}
[{\includegraphics[width=1in,height=1.25in,clip,keepaspectratio]{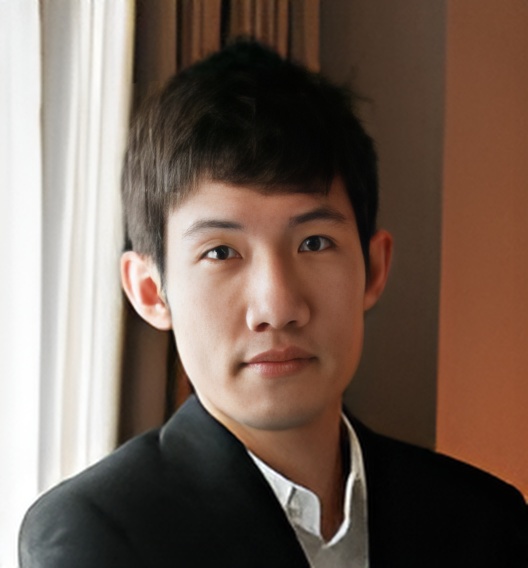}}]{Hong-Han Shuai} received the B.S. degree from the Department of Electrical Engineering, National Taiwan University (NTU), Taipei, Taiwan, R.O.C., in 2007, the M.S. degree in computer science from NTU in 2009, and the Ph.D. degree from Graduate Institute of Communication Engineering, NTU, in 2015. He is currently an Associate Professor at National Yang Ming Chiao Tung University (NYCU). His research interests are in the areas of multimedia processing, machine learning, social network analysis, and data mining. His works have appeared in top-tier conferences such as MM, CVPR, AAAI, KDD, WWW, ICDM, CIKM, and VLDB, and top-tier journals such as TKDE, TMM, TNNLS, and JIOT. Moreover, he has served as a PC member for international conferences, including MM, AAAI, IJCAI, and WWW, and the invited reviewer for journals including TKDE, TMM, JVCI, and JIOT.
\end{IEEEbiography}

\begin{IEEEbiography}
[{\includegraphics[width=1in,height=1.25in, clip,keepaspectratio]{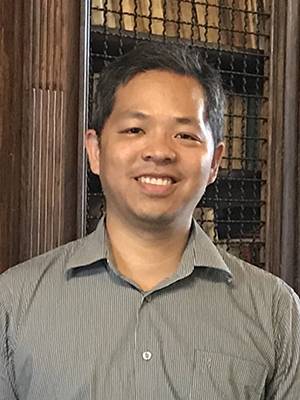}}]{Wen-Huang Cheng} is a Professor at the Department of Computer Science and Information Engineering, National Taiwan University (NTU). His current research interests include artificial intelligence, multimedia, computer vision, machine learning, digital transformation, and financial technology. He has actively participated in international events and played important leading roles in prestigious journals, conferences, and professional organizations, including Editor-in-Chief for IEEE CTSoc News on Consumer Technology, Senior Editor for IEEE Consumer Electronics Magazine, Associate Editor for IEEE Transactions on Multimedia (TMM), General Chair for ACM MMAsia (2023), IEEE ICCETW (2023, 2022), IEEE ICME (2022) and ACM ICMR (2021), Chair for IEEE Multimedia Systems and Applications (MSA) technical committee, governing board member for IAPR. He has received numerous research and service awards, including the Best Paper Award of 2021 IEEE ICME and the Outstanding Associate Editor Award of IEEE T-MM (2021 and 2020, twice). He is an IEEE Distinguished Lecturer, ACM Distinguished Member, and IET Fellow.
\end{IEEEbiography}

\end{document}